%% file: main.tex
\documentclass[letterpaper,11pt]{article}
\usepackage[margin=1in]{geometry}
\usepackage[utf8]{inputenc}
\usepackage[T1]{fontenc}
\usepackage{libertine}
\usepackage{amsmath,amssymb}
\usepackage{libertinust1math}
\usepackage{macros}
\hypersetup{
    pdftitle={The Optimal Sample Complexity of Agnostic PAC Learning for Partial Concept Classes}
}

\title{\LARGE\bfseries The Optimal Sample Complexity of Agnostic PAC Learning\allowbreak{} for Partial Concept Classes}

\author{
    \begin{minipage}[t]{0.45\textwidth}
        \centering
        {\normalsize Liang-Yu Zou} \\
        {\small \textit{Aarhus University}} \\
        {\small$\mathtt{zou@cs.au.dk}$}
    \end{minipage}
}

\date{}

\begin{document}
\maketitle
\input{content/0.abstract.tex}
\input{content/1.introduction.tex}
\input{content/3.generalized_rademacher.tex}
\input{content/4.agnostic_learner.tex}

\section*{AI Disclosure}
In mid-October, ChatGPT-6 Astra produced an initial draft of this paper. 
The authors subsequently became aware of the new manuscript \cite{alon2026uniform} and then asked ChatGPT-6 Astra to simplify the draft using the $\eps$-disambiguation technique.


\begingroup
\raggedright
\bibliographystyle{alpha}
\bibliography{reference}
\endgroup

\appendix

\input{content/a.missing_proof_grad.tex}

\input{content/b.missing_proof_pac.tex}

\end{document}

%% file: content/0.abstract.tex
\begin{abstract}
    Let $\mathcal H\subseteq\{-1,+1,\star\}^{\mathcal X}$ be a partial concept class of finite VC dimension $d$. We construct a deterministic agnostic PAC learner that, for every $\varepsilon,\delta\in(0,1)$, uses $O((d+\log(1/\delta))/\varepsilon^2)$ i.i.d. examples from an arbitrary distribution $\mathcal D$ over $\mathcal X\times\{-1,+1\}$ and returns a total concept $\hat h:\mathcal X\to\{-1,+1\}$ satisfying $\text{er}_{\mathcal D}(\hat h)\leq \min_{h\in\mathcal H}\text{er}_{\mathcal D}(h)+\varepsilon$ with probability at least $1-\delta$, where $\text{er}_{\mathcal D}(h)=\mathbb P_{(x,y)\sim\mathcal D}[h(x)\neq y]$. This establishes the optimal sample complexity, matching the lower bound of Alon et al. (FOCS 2021).
\end{abstract}

%% file: content/1.introduction.tex
\section{Introduction}

In the classical PAC model~\cite{DBLP:journals/cacm/Valiant84},
a total concept assigns a binary label to every point of an
instance space $\cX$.
\emph{Partial concept classes}, introduced in \cite{DBLP:conf/focs/AlonHHM21}, relax this requirement: a partial concept is a function $h:\cX\to\{-1,+1,\star\}$, and $h(x)=\star$ means that $h$ is undefined at $x$.
This framework can capture many natural assumptions on the data, such as the existence of a large-margin region around a halfspace boundary on which the concept is undefined.

The VC dimension \cite{vapnik1974theory} extends naturally to partial concept classes.
A finite set $\bm x=\{x_1,\ldots,x_n\}\subseteq\cX$ is shattered by $\cH\subseteq\{-1,+1,\star\}^{\cX}$ if every binary labeling of $\bm x$ is realized by a concept in $\cH$.
The VC dimension $\VC(\cH)$ is the largest cardinality of a shattered set, or infinity if arbitrarily large finite sets are shattered.
As in the classical setting, finite VC dimension characterizes learnability~\cite{DBLP:journals/jacm/BlumerEHW89, DBLP:conf/focs/AlonHHM21}.

In the realizable case, the data is labeled by a concept in the class, a PAC learner is required to output a concept such that with high probability $1-\delta$, the output concept has expected error at most $\eps$.
For both total concept classes and partial concept classes, the optimal sample complexity in this setting is $\Theta((d+\log(1/\delta))/\eps)$, where $d$ is the VC dimension of the concept class \cite{DBLP:journals/jmlr/Hanneke16,DBLP:conf/focs/AlonHHM21,DBLP:conf/colt/Larsen23,DBLP:conf/focs/Aden-AliCSZ23,DBLP:journals/corr/abs-2606-13614}.

We study agnostic PAC learning \cite{DBLP:journals/ml/KearnsSS94}, in which the data distribution is arbitrary and the learner competes with the best concept in a prescribed concept class.
For a distribution $\cD$ over $\cX\times\{-1,+1\}$ and a labeled sample $S=((x_i,y_i))_{i=1}^n$ i.i.d. from $\cD$, define the expected and empirical error by $\er_\cD(h) =\Prob_{(x,y)\sim\cD}[h(x)\neq y]$ and $\erhat_S(h)=\frac1n\sum_{i=1}^n\ind_{\{h(x_i)\neq y_i\}}$ respectively.
In particular, an undefined prediction $\star$ always incurs a misclassification.
For a nonempty class $\cH$, let $\ermin_\cD(\cH)=\min_{h\in\cH}\er_\cD(h)$.\footnote{We assume that all functions of the data are measurable and that this minimum is attained.}
The goal of agnostic PAC learning is to output a hypothesis $\hat h$ such that $\er_\cD(\hat h)\leq \ermin_\cD(\cH)+\eps$ with probability at least $1-\delta$, for given accuracy $\eps\in (0,1)$ and failure probability $\delta\in (0,1)$.
And we define the sample complexity of a learning algorithm as the minimum number of samples required to achieve this goal.%
\footnote{\cite{DBLP:conf/focs/AlonHHM21} introduces an alternative definition of agnostic PAC learning, and we show that our algorithm also satisfies this alternative definition. See \cref{sec:alter-agnostic} for details.}

For total concept classes, the optimal sample complexity of agnostic PAC learning is $\Theta((d+\log(1/\delta))/\eps^2)$ \cite{talagrand1994sharper}, through the empirical risk minimization (ERM) algorithm, using uniform convergence arguments.
Very recently, \cite{mathiasen2026optimal} recovers the same sample complexity for total concept classes using a different algorithm based on aggregation of agnostic one-inclusion graph algorithms, building on the work of \cite{DBLP:journals/ml/Long99,DBLP:conf/colt/AsilisDDST24,DBLP:conf/alt/DughmiKY25}.
For partial concept classes, \cite{DBLP:conf/focs/AlonHHM21} shows that the ERM algorithm is not guaranteed to work, and establishes the lower bound of $\Omega((d+\log(1/\delta))/\eps^2)$, matching the total concept class case, and the state-of-the-art upper bound of
\[
    O\left(\frac{d\log^2(d/\eps)+\log(1/\delta)}{\eps^2}\right),
\]
leaving a gap of $\log^2(d/\eps)$ between the lower and upper bounds.

\subsection{Main Results}
We close the logarithmic gap for agnostic PAC learning of partial concept classes, and show that the optimal sample complexity of agnostic PAC learning for partial concept classes is $\Theta((d+\log(1/\delta))/\eps^2)$, matching the total concept class case.
\begin{theorem}[Informal \cref{thm:pac}]
    There exists an agnostic PAC learning algorithm, \cref{alg:pac}, for partial concept classes with VC dimension $d\geq 1$ that uses $O((d+\log(1/\delta))/\eps^2)$ samples and outputs a hypothesis such that with probability at least $1-\delta$, $\er_\cD(\hat h)\leq \min_{h\in\cH}\er_\cD(h)+\eps$.
\end{theorem}
To achieve this result, we generalize the Rademacher complexity to partial concept classes, and bound it using the chaining technique \cite{dudley1967sizes,DBLP:journals/jct/Haussler95}.
\begin{definition}[Generalized Rademacher complexity]
    For a partial concept class $\cH\subseteq\{-1,+1,\star\}^n$, the generalized Rademacher complexity is defined as
    \[
        \gRad(\cH)=\frac{1}{n}\E_{\bm\sigma}\left[\max_{h\in\cH}\sum_{i\in[n]}\sigma_i h_i\right],
    \] 
    where $\bm\sigma=(\sigma_1,\ldots,\sigma_n)$ is a vector of i.i.d. Rademacher random variables, i.e., $\Prob(\sigma_i=1)=\Prob(\sigma_i=-1)=1/2$. The product between $\star$ and any value in $\{-1,+1\}$ is defined as $-1$. 
\end{definition}
When $\cH$ is a total concept class, the generalized Rademacher complexity degenerates to the Rademacher complexity.
The main challenge is that the only known approach to cover partial concept classes is to use the $\eps$-disambiguations \cite{alon2026uniform}.
However, the $\eps$-disambiguation only considers the defined coordinates, thus a lot of undefined coordinates may incur large Euclidean distance between two disambiguations, which makes the chaining technique fail.
To address this issue, we use these disambiguations to construct real-valued vectors to approximate the partial concepts with different scales.
The construction controls the Euclidean distance between two vectors and the number of different approximation vectors, which allows us to apply the chaining technique.
\begin{theorem}[Informal \cref{thm:grad}]
    Let $n\geq 1$ and let $\cH\subseteq\{-1,+1,\star\}^n$ be a partial concept class with VC dimension $d\geq 0$. Then, it holds that $\gRad(\cH)\leq C\sqrt{d/n}$ for a universal constant $C>0$.
\end{theorem}

With this in hand, we convert the generalized Rademacher complexity bound to a leave-one-out bound, and then a PAC bound. 
We provide a leave-one-out learner by adapting the minimax forecaster construction \cite{DBLP:conf/birthday/Cesa-BianchiS13} to obtain real-valued predictions satisfying the leave-one-out bound, without using the agnostic one-inclusion graph algorithm, which may be of independent interest.
Finally, we adapt the aggregation technique of \cite{mathiasen2026optimal} to convert the leave-one-out learner to a PAC learner, using a thresholding step to convert the real-valued predictions to binary predictions.

%% file: content/3.generalized_rademacher.tex
\section{Bounding the generalized Rademacher complexity}
\label{sec:grad}
Recall the definition of the generalized Rademacher complexity of a partial concept class $\cH\subseteq\{-1,+1,\star\}^n$:
\[
    \gRad(\cH)
    =\frac{1}{n}\E_{\bm\sigma}\left[\max_{h\in\cH}\sum_{i\in[n]}\sigma_i h_i\right],
\] 
where the products $+1\cdot \star$ and $-1\cdot \star$ are both defined as $-1$.
We now show that the generalized Rademacher complexity of a partial concept class $\cH$ with VC dimension $d$ is upper bounded by $O(\sqrt{d/n})$.
Throughout this section, we write $h_i$ as the $i$-th coordinate of $h\in\cH$ for simplicity.

\begin{theorem}
    \label{thm:grad}
    Let $n\geq 1$ and let $\cH\subseteq\{-1,+1,\star\}^n$ be a partial concept class with VC dimension $d\geq 0$. There is a universal constant $C>0$ such that
    \[
        \gRad(\cH)\leq C\sqrt{d/n}.
    \] 
\end{theorem}
Before presenting the proof, we remark that the generalized Rademacher complexity bound is not enough to guarantee uniform convergence for partial concept classes.
In standard proofs of uniform convergence, introducing a ghost sequence $S'$ and applying the symmetrization technique give the following inequality (see \cite[Pages 31-34]{10.7551/mitpress/10290.001.0001} for example):
\[
    \E_{S}\left[\sup_{h}\left(\hat\er_{S}(h)-\er_\cD(h)\right)\right]
    \leq\frac{2}{n}\E_{\bm\sigma,S}\left[\sup_{h\in\cH}\sum_{i=1}^n\sigma_i\ind_{\{h(x_i)\neq y_i\}}\right]
    =\frac{1}{n}\E_{\bm\sigma,S}\left[\sup_{h\in\cH}\sum_{i=1}^n-\sigma_i\cdot(y_ih(x_i))\right],
\]
where the last equality holds because $\ind_{\{h(x_i)\neq y_i\}}=\frac{1-y_ih(x_i)}{2}$.
For total concept classes, the last term is connected to Rademacher complexity by the fact that $\E_{\bm\sigma}[\sup_{h\in\cH}\sum_{i=1}^n-\sigma_i\cdot(y_ih(x_i))]=\E_{\bm\sigma}[\sup_{h\in\cH}\sum_{i=1}^n\sigma_i h(x_i)]$ over the randomness of $\bm\sigma$.
However, for partial concept classes, this equality does not hold because when $h(x_i)=\star$, we have $-\sigma_i\cdot(y_i\star)=\sigma_i$ but $\sigma_i \star=-1$.

The proof follows the classical chaining technique, which is a standard method to bound the Rademacher complexity of a total concept class \cite{dudley1967sizes,DBLP:journals/jct/Haussler95}. 
The obstacle is that, for a partial concept class, we don't have a direct way to construct a sequence of total concepts which approximates the partial concept with different scales, and the Euclidean distance between two neighboring approximations is bounded by the scale.
We overcome this obstacle by first constructing a sequence of total concepts which approximately disambiguates the partial concept with different scales, and then using these disambiguations to construct a sequence of $[-1,1]^n$ vectors which approximate the partial concept with different scales and the euclidean distance between them is bounded by the scale.

For every $\bm\sigma$, let $h_{\bm\sigma}$ be the maximizer of its corresponding term in the definition of $\gRad(\cH)$, i.e.,
\[
    h_{\bm\sigma}=\argmax_{h\in\cH}\sum_{i\in[n]}\sigma_i h_{i},
\] 
break ties using a fixed rule. Thus,
\[
    \gRad(\cH)
    =\frac{1}{n}\E_{\bm\sigma}\left[\sum_{i\in[n]}\sigma_i h_{\bm\sigma,i}\right]
    =\frac{1}{n}\E_{\bm\sigma}\left[\langle \bm\sigma, h_{\bm\sigma} \rangle\right].
\] 
As in the classical chaining technique, we want to construct a sequence of approximations of $h_{\bm\sigma}$ with different scales, and write the sum of $\sum_{i\in[n]}\sigma_i h_{\bm\sigma}$ as a telescoping sum of the approximations.
Let 
\[
    J=\lfloor \log_2n \rfloor+1
\] 
be the number of telescoping steps. For every $h_{\bm\sigma}$, let $\{\bg_{\bm\sigma}^j\}_{j=0}^J$ be a sequence of approximations, precision increasing with $j$, and set the raw approximation $\bg_{\bm\sigma}^0$ to be a constant vector for all $\bm\sigma$. 
Suppose the terminal approximation $\bg_{\bm\sigma}^J$ upper bounds $h_{\bm\sigma}$ in terms of the expected product with the Rademacher variables, we use a telescoping sum of the approximations to bound the original sum:
\begin{align*}
    \E_{\bm\sigma}\left[\langle \bm\sigma, h_{\bm\sigma} \rangle\right]
    \leq\E_{\bm\sigma}\left[\langle \bm\sigma, \bg_{\bm\sigma}^J\rangle\right]    
    =\E_{\bm\sigma}\left[\langle \bm\sigma, \bg_{\bm\sigma}^0 \rangle+\sum_{j=1}^J \langle \bm\sigma, \bg_{\bm\sigma}^j-\bg_{\bm\sigma}^{j-1}\rangle\right]
    =\E_{\bm\sigma}\left[\sum_{j=1}^J \langle \bm\sigma, \bg_{\bm\sigma}^j-\bg_{\bm\sigma}^{j-1}\rangle\right],
\end{align*} 
which implies that 
\begin{align}
    \label{eq:grad-telescope}
    \gRad(\cH)
    &\leq\frac{1}{n} \E_{\bm\sigma}\left[\sum_{j=1}^J\langle \bm\sigma,  \bg_{\bm\sigma}^j-\bg_{\bm\sigma}^{j-1}\rangle\right]
\end{align}
To use the chaining technique, the requirement for these approximations is that, in expectation over $\bm\sigma$, $\langle \bm\sigma, \bg_{\bm\sigma}^J\rangle$ bounds $\langle \bm\sigma, h_{\bm\sigma} \rangle$, $\bg_{\bm\sigma}^0$ is a constant vector for all $\bm\sigma$, every increment $\norm{\bg_{\bm\sigma}^j-\bg_{\bm\sigma}^{j-1}}_2^2$ is bounded by the scale of the approximation, and the number of possible increments is bounded.
\subsection{Limitations of \texorpdfstring{$\eps$}{epsilon}-disambiguation}
A natural way to construct these approximations is to use total concepts. \cite{alon2026uniform} introduced the notion of $\eps$-disambiguation and constructed size-bounded $\eps$-disambiguations of partial concept classes.
For two partial concepts $h,h'\in\{-1,+1,\star\}^n$, define the \textit{normalized conflict} between them as
\[
    c(h,h')=\frac{1}{n}\left| \{i\in[n]: h_i\neq \star, h'_i\neq \star, h_i\neq h'_i\} \right|,
\]
which is the fraction of coordinates where both $h$ and $h'$ are defined but have opposite values. 
\begin{definition}[$\eps$-disambiguation]
    \label{def:disambiguation}
    Let $\eps\in (0,1/2]$, $n\geq 1$, and let $\cH\subseteq\{-1,+1,\star\}^n$ be a partial concept class with VC dimension $d\geq 1$, an $\eps$-disambiguation of $\cH$ is a total concept class $\widebar\cH\subseteq\{-1,+1\}^n$ such that for every $h\in \cH$, there exists $f\in\widebar\cH$ such that $c(h,f)\leq \eps$.
\end{definition}
\begin{lemma}[{\cite[Theorem 2.4]{alon2026uniform}}]
    \label{lem:eps-disambiguation}
    Let $\eps\in (0,1/2]$, $n\geq 1$, and let $\cH\subseteq\{-1,+1,\star\}^n$ be a partial concept class with VC dimension $d\geq 1$, there exists an $\eps$-disambiguation $\widebar\cH$ of $\cH$ such that
    \[
        |\widebar\cH|\leq (1/\eps)^{C d\log(1/\eps)},
    \]
    where $C>0$ is a universal constant.
\end{lemma}
With this lemma, we can construct a sequence of $\eps$-disambiguations $\widebar\cH^1,\ldots, \widebar\cH^J$ of $\cH$ with different scales $2^{-j}$ for $1\leq j\leq J$, where $J=\lfloor \log_2 n\rfloor+1$.
However, these disambiguations cannot be used directly as the approximations, because the euclidean distance between two neighboring disambiguations is not bounded by the scale of the disambiguation.
Let $\widebar h^j, \widebar h^{j-1}$ be the disambiguations of a partial concept $h$ at scales $2^{-j}$ and $2^{-j+1}$, respectively, then we have
\[
    c(h,\widebar h^j)\leq 2^{-j} ,\quad c(h,\widebar h^{j-1})\leq 2^{-j+1} ,
\]
but the conflict only counts the differences on defined coordinates of $h$, and the number of defined coordinates of $h$ can be as small as $0$, which means that the squared euclidean distance between $\widebar h^j$ and $\widebar h^{j-1}$ can be as large as $\Theta(n)$, which is much larger than the scale $2^{-j}$.

\subsection{Real-valued approximations}
Knowing that the $\eps$-disambiguation is not enough, we relax the approximations from total concepts to vectors in $[-1,1]^n$.
Let $\bg_{\bm\sigma}=(g_{\bm\sigma,1},\ldots,g_{\bm\sigma,n})$ be a vector in $[-1,1]^n$ which approximates a partial concept $h_{\bm\sigma}$.
Consider the approximation error $\langle \bm\sigma, h_{\bm\sigma}\rangle-\langle \bm\sigma, \bg_{\bm\sigma} \rangle$, 
on a defined coordinate $i$ of $h_{\bm\sigma}$, the approximation error is $\sigma_i (h_{\bm\sigma,i}-g_{\bm\sigma,i})\leq |h_{\bm\sigma,i}-g_{\bm\sigma,i}|$; on an undefined coordinate, the approximation error is $-1-\sigma_i g_{\bm\sigma,i}\leq |\sigma_i g_{\bm\sigma,i}|-1$. 
We thus define the \textit{approximation error} as follows: let $g\in [-1,1]$ be an approximation of a label $y\in \{-1,+1,\star\}$, the approximation error is defined as
\begin{align*}
    \rho(g,y) = \begin{cases}
        |g-y|, & y\in\{-1,+1\}, \\
        |g|-1, & y=\star.
    \end{cases}
\end{align*}
Then, we have 
\begin{align*}
    \E_{\bm\sigma}\left[\langle \bm\sigma, h_{\bm\sigma} \rangle\right]
    \leq\E_{\bm\sigma}\left[\langle \bm\sigma, \bg_{\bm\sigma}^J\rangle+\sum_{i\in[n]} \rho(g_{\bm\sigma,i}^J, h_{\bm\sigma,i})\right].
\end{align*}
Substituting the above relation into \cref{eq:grad-telescope} gives
\[
    \gRad(\cH)
    \leq \frac{1}{n}\left(\E_{\bm\sigma}\left[\sum_{j=1}^J \langle \bm\sigma, \bg_{\bm\sigma}^j-\bg_{\bm\sigma}^{j-1}\rangle\right]+\E_{\bm\sigma}\left[\sum_{i\in[n]} \rho(g_{\bm\sigma,i}^J, h_{\bm\sigma,i})\right]\right).
\]
If the expected terminal approximation error is nonpositive, then we can ignore the terminal approximation error and only need to consider the increment terms as the classical chaining technique.   
Let 
\[
    I^j=\{\bg_{\bm\sigma}^j-\bg_{\bm\sigma}^{j-1}: \bm\sigma\in\{\pm 1\}^n\}
\]
be the set of all possible increment vectors at scale $j$, we need to bound the size of $|I^j|$ and the increments $\norm{\bg_{\bm\sigma}^j-\bg_{\bm\sigma}^{j-1}}_2^2$ for every $j$ to use the chaining technique.
The following lemma shows that such a sequence of approximations exists.

\begin{lemma}
    \label{lem:grad-approximate-vector}
    Let $\cH\subseteq\{-1,+1,\star\}^n$ be a partial concept class with VC dimension $d\geq 1$, $J=\lfloor \log_2n\rfloor+1$.
    For every Rademacher vector $\bm\sigma$, let $h_{\bm\sigma}$ be the maximizer of $\sum_{i\in[n]}\sigma_i h_{\bm\sigma,i}$, there exists a sequence of approximation vectors $\{\bg^j_{\bm\sigma}\}_{j=0}^J$ with $\bg^j_{\bm\sigma}\in [-1,1]^n$ for $0\leq j\leq J$, such that 
    \begin{itemize}
        \item $\bg^0_{\bm\sigma}$ is a constant vector for all $\bm\sigma$;
        \item $\norm{\bg_{\bm\sigma}^j-\bg_{\bm\sigma}^{j-1}}_2^2\leq 256n\cdot 2^{-j}$ for all $\bm\sigma$ and all $1\leq j\leq J$;
        \item $\E_{\bm\sigma}\left[\sum_{i\in[n]} \rho(g_{\bm\sigma,i}^J, h_{\bm\sigma,i})\right]\leq 0$;
        \item $|I^j|\leq \exp(Cdj^3)$ for all $1\leq j\leq J$, where $C>0$ is a universal constant. 
    \end{itemize}
\end{lemma}
The proof of this lemma is given in \cref{sec:proof-approximate-vector}. 
With this lemma, we can apply the chaining technique to prove the generalized Rademacher complexity bound in \cref{thm:grad}, the proof of which is given in \cref{sec:proof-chaining}.

We now show how to construct the sequence of approximation vectors $\{\bg^j_{\bm\sigma}\}_{j=0}^J$ for every $\bm\sigma$.
For $1\leq j\leq J$, construct a sequence of $\eps$-disambiguations $\widebar \cH^1,\ldots, \widebar \cH^J$ of $\cH$ with different scales $2^{-j}$.  
With these disambiguations fixed, for every $\bm\sigma$, let $h_{\bm\sigma}$ be the maximizer of $\langle\bm\sigma, h\rangle$ over $h\in\cH$, and let $\widebar h_{\bm\sigma}^j$ be the $2^{-j}$-disambiguation of $h_{\bm\sigma}$ for $1\leq j\leq J$, breaking ties according to a fixed rule.
This information is thus fixed for all $\bm\sigma$ before we construct the approximation of a specific $h_{\bm\sigma}$.
To construct the approximation of $h_{\bm\sigma}$, we only use the information of the disambiguations $\widebar h_{\bm\sigma}^1,\ldots, \widebar h_{\bm\sigma}^J$ of $h_{\bm\sigma}$, and we don't need to know $h_{\bm\sigma}$ and $\bm\sigma$ themselves.
\paragraph{Coordinate-wise construction.}
Fix $\bm\sigma$, then $h_{\bm\sigma}$ and $\{\widebar h_{\bm\sigma}^j\}_{j=1}^J$ are also fixed.
Assume that we don't know $\bm\sigma$, let $(\widebar h^1,\ldots, \widebar h^J)$ be the observed sequence of $\eps$-disambiguations.
Let $\{\cF^j\}_{j=0}^J$ be a filtration of information, where $\cF^j$ is the information generated by the prefix $(\widebar h^1,\ldots, \widebar h^j)$ and $\cF^0$ is trivial, containing the fixed information.

Fix a coordinate $i\in [n]$, we now construct a sequence of approximation values $\{g_{i}^j\}_{j=0}^J$ for this coordinate. 
Let $Y\in\{-1,+1,\star\}$ be a random variable used to guess the value of $h_{\bm\sigma,i}$.
When seeing the information $\cF^j$, let $\Sigma^j$ be the set of Rademacher vectors $\bm\sigma$ whose maximizer $h_{\bm\sigma}$ has the $\eps$-disambiguations as the prefix $(\widebar h^1,\ldots, \widebar h^j)$, i.e.,
\[
    \Sigma^j=\{\bm\sigma\in\{\pm 1\}^n: \widebar h_{\bm\sigma}^k=\widebar h^k \text{ for } 1\leq k\leq j\},
\]
then the conditional probability that $Y=+1$ is defined as the fraction of Rademacher vectors in $\Sigma^j$ whose maximizer $h_{\bm\sigma}$ has $h_{\bm\sigma,i}=+1$, i.e.,
\[
    \Prob[Y=+1|\cF^j]
    =\frac{\left|\bm\sigma\in \Sigma^j: h_{\bm\sigma,i}=+1\right|}{\left|\Sigma^j\right|},
\]
and similarly for $Y=-1$ and $Y=\star$. For convenience,
define
\[
    p^j=\Prob[Y=+1|\cF^j],
    \quad q^j=\Prob[Y=-1|\cF^j],
    \quad v^j=p^j+q^j=\Prob[Y\neq \star|\cF^j].
\]
Since $J=\lfloor \log_2n\rfloor+1$, we have $2^{-J}< 1/n$, the definition of $\eps$-disambiguation implies that, if there are two partial concepts $h,h'\in\cH$ whose $2^{-J}$-disambiguations are the same $\widebar h^J$, then 
\[
    c(h,\widebar h^J)< 1/n,\quad
    c(h',\widebar h^J)< 1/n,
\]
which means that on every coordinate where both $h$ and $h'$ are defined, they have the same value as $\widebar h^J$.
Thus, one of $\{\bm\sigma\in \Sigma^J: h_{\bm\sigma,i}=+1\}$ and $\{\bm\sigma\in \Sigma^J: h_{\bm\sigma,i}=-1\}$ is empty, which implies that $p^J$ and $q^J$ cannot be both positive, i.e., $p^J q^J=0$.

The general idea is to use the conditional mean of $Y$ conditioned on the information $\cF^j$ and the event that $Y$ is defined as the approximation value $g_i^j$.
Define the conditional mean as
\begin{align*}
    \theta^j=
        \begin{cases}
            (p^j-q^j)/v^j, & v^j>0,\\
            0, & v^j=0.
        \end{cases}
\end{align*}
The range of $\theta^j$ is $[-1,1]$. 
Since $p^J q^J=0$, if $v^J>0$, then $\theta^J\in\{-1,+1\}$ and $\theta^J=h_{\bm\sigma,i}$ when $h_{\bm\sigma,i}$ is defined, thus $\rho(\theta^J, h_{\bm\sigma,i})\leq 0$ no matter whether $h_{\bm\sigma,i}$ is defined or not; if $v^J=0$, then $\theta^J=0$ and $h_{\bm\sigma,i}$ must be undefined by the definition of $v^J$, thus $\rho(\theta^J, h_{\bm\sigma,i})=|0|-1=-1<0$. 
Therefore, the terminal approximation error $\rho(\theta^J, h_{\bm\sigma,i})$ is always nonpositive.

However, directly using $\theta^j$ as the approximation can lead to large increments. 
When $v^{j-1},v^j>0$, the increment satisfies
\[
    |\theta^j-\theta^{j-1}|^2
    =\frac{4(p^{j-1}q^j-p^jq^{j-1})^2}{(v^{j-1}v^j)^2}
    \leq \frac{4(p^{j-1}q^j+p^jq^{j-1})}{v^{j-1}v^j},
\]
whose numerator $p^{j-1}q^j+p^jq^{j-1}$ is closely related to the approximation scale, which will be shown later.
If $v^{j-1}$ or $v^j$ is small, the denominator can be very small and the increment can be out of control. 
To avoid this problem, when we see a $v^j$ smaller than a constant threshold, we use a convex combination of $\theta^j$ and the previous approximation $g^{j-1}$ as the new approximation $g^j$, which is closer to $g^{j-1}$ and thus has a smaller increment.
We also set every remaining approximation $g^k$ with $j< k\leq J$ to be the same value as $g^j$, which surprisingly still guarantees that the terminal approximation error is nonpositive in conditional expectation given the prefix at which it stops.

The coordinate-wise construction is given as follows.
Run the following construction for every coordinate $i\in [n]$ independently, then the sequence of approximation vectors $\{\bg^j_{\bm\sigma}\}_{j=0}^J$ is constructed by combining the expected approximations of all coordinates.

\begin{tcolorbox}[colback=gray!10,colframe=gray!10,boxrule=0pt,arc=0pt]
    If $v^0\leq 1/4$, set $g^j=0$ for all $j$ and stop; otherwise set $g^0=\theta^0$ and continue.
    For $j=1,\ldots, J$, if $v^j>1/4$, set $g^j=\theta^j$ and continue; otherwise, set
    \[
        g^j=(1-2v^j)g^{j-1}+2v^j\theta^j,
    \]
    and set every remaining $g^k$ with $j< k\leq J$ to be the same value as $g^j$ and stop.
\end{tcolorbox}
 

%% file: content/4.agnostic_learner.tex
\section{Agnostic PAC learning with optimal sample complexity}
\label{sec:pac}
In this section, we present an agnostic PAC learner for partial concept classes that achieves the optimal sample complexity. 
Actually, plugging the generalized Rademacher complexity bound in \cref{thm:grad} into the analysis of the agnostic one-inclusion graph algorithm in \cite{DBLP:conf/alt/DughmiKY25} gives us the required leave-one-out bound. 
To convert the leave-one-out bound to a PAC bound, \cite{mathiasen2026optimal} uses the predictions of many agnostic one-inclusion graph algorithms to construct real-valued predictions, and then aggregates these predictions to construct a PAC learner.

The novelty of this section is that we adapt the minimax forecaster construction \cite{DBLP:conf/birthday/Cesa-BianchiS13} to obtain real-valued predictions satisfying the leave-one-out bound, without using the agnostic one-inclusion graph algorithm.
We then adapt the aggregation technique of \cite{mathiasen2026optimal} to convert this bound into a PAC bound.

\subsection{Real-valued prediction and leave-one-out bound}
Throughout this section, fix a partial concept class $\cH\subseteq\{-1,+1,\star\}^\cX$ with VC dimension $d<\infty$.
For a labeled sequence $S=((x_j,y_j))_{j=1}^n$, let $S_{-i}$ denote the sequence $S$ with the $i$-th example removed.
Let $\cH_{\bm x}=\{(h(x_1),\ldots,h(x_n)):h\in\cH\}\subseteq\{-1,+1,\star\}^n$ be the restriction of $\cH$ to the unlabeled sequence $\bm x=(x_1,\ldots,x_n)$.
For ease of presentation, define the unnormalized minimum empirical risk on a labeled sequence $T$ with respect to the hypothesis class $\cH$ as
\[
    N(T)=\min_{h\in\cH}\sum_{(x,y)\in T}\ind_{\{h(x)\ne y\}},
\] 
which is the minimum number of misclassified examples by any hypothesis in $\cH$ on the sequence $T$, and it satisfies $N(T)=|T|\cdot\erhatmin_T=|T|\cdot \min_{h\in\cH}\erhat_T(h)$.
Besides, define the loss between a real value $a\in[-1,1]$ and a label $b\in\{-1,+1\}$ as
\[
    \ell(a,b)=\frac{|a-b|}{2}=\frac{1-ab}{2},
\]
which generalizes the $0$-$1$ loss between two labels in $\{-1,+1\}$.
Our goal is to construct a leave-one-out learner satisfying the following lemma.

\begin{lemma}[Leave-one-out bound]
    \label{lem:leave-one-out}
    For every labeled sequence $S=((x_j,y_j))_{j=1}^n$ of size $n\geq 1$ with $\bm x=(x_1,\ldots,x_n)$, there exists a leave-one-out learner, \cref{alg:leave-one-out}, which, for every $i\in[n]$, takes  the sequence $S_{-i}$ and the query $x_i$ as input, and outputs a real-valued prediction $\mu_{S_{-i}}(x_i)\in[-1,1]$ such that
    \[
        \frac{1}{n}\sum_{i=1}^n\ell(\mu_{S_{-i}}(x_i),y_i)
        =\erhatmin_S(\cH)+\frac{\gRad(\cH_{\bm x})}{2}
        \leq \erhatmin_S(\cH)+C\sqrt{\frac{d}{n}},
    \]
    where $C>0$ is a universal constant and $\erhatmin_S(\cH)=\min_{h\in\mathcal H}\erhat_S(h)$ is the minimum empirical risk.
\end{lemma}

The leave-one-out learner is motivated by the minimax forecaster of \cite[Algorithm 1]{DBLP:conf/birthday/Cesa-BianchiS13}. 
That forecaster retains the previously observed labels, assigns independent uniform signs to future labels, and predicts using the expected difference between the empirical optima obtained by assigning the current query the labels $-1$ and $+1$.

Our learner instead averages the difference between the empirical optima obtained by assigning the query $x_i$ the labels $-1$ and $+1$ over corrupted training sequences $\widetilde S_{-i}$. 
To generate each corrupted sequence, we draw $\beta\sim\Unif[0,1/2]$ and, conditional on $\beta$, independently flip each training label with probability $\beta$.

The intuition comes from considering the expected empirical optimum when the full sequence $S$ is corrupted in the same way.
At $\beta=0$, the sequence is unchanged, so this expectation equals the empirical optimum on the original sequence, i.e., $\erhatmin_S(\cH)$.
At $\beta=1/2$, the corrupted labels are independent uniform signs, and the expected empirical optimum can be expressed through the generalized Rademacher complexity, i.e. $\gRad(\cH_{\bm x})$. 
The differences, which define our predictions, form the derivative of this expected empirical optimum with respect to $\beta$. 
Integrating from $0$ to $1/2$ therefore connects the leave-one-out loss to these two endpoint quantities, yielding the desired bound.

We now describe the theoretical construction of the leave-one-out learner, while the implementation is given in \cref{alg:leave-one-out}.

\begin{tcolorbox}[colback=gray!10,colframe=gray!10,boxrule=0pt,arc=0pt]
    The prediction is given by $\mu_{S_{-i}}(x_i)=\E_{\beta,\widetilde S_{-i}}[\nu_{\widetilde S_{-i}}(x_i)]$, where the expectation is taken over the corruption rate $\beta\sim \Unif[0,1/2]$ and the corrupted sequence $\widetilde S_{-i}$. 
    One realization of $\nu_{\widetilde S_{-i}}(x_i)$ is obtained by the following procedure:
    \begin{enumerate}
        \item Draw a corruption rate $\beta\sim \Unif[0,1/2]$;
        \item For each example $(x_j,y_j)$ in $S_{-i}$, flip the label $y_j$ independently with probability $\beta$ to obtain a corrupted label $\widetilde y_j$, and let $\widetilde S_{-i}=((x_j,\widetilde y_j))_{j\in[n]\setminus\{i\}}$ be the corrupted sequence;
        \item Let $\widetilde S^{+1}$ be the corrupted full sequence obtained by adding $(x_i,+1)$ to $\widetilde S_{-i}$ at coordinate $i$, similarly define $\widetilde S^{-1}$ by adding $(x_i,-1)$ to $\widetilde S_{-i}$;
        \item Set $\nu_{\widetilde S_{-i}}(x_i)=N(\widetilde S^{-1})-N(\widetilde S^{+1})$.
    \end{enumerate} 

\end{tcolorbox}
Note that $N(\widetilde S_{-i})\leq N(\widetilde S^{-1}),N(\widetilde S^{+1})\leq N(\widetilde S_{-i})+1$, $\nu_{\widetilde S_{-i}}(x_i)$ must lie in $\{-1,0,+1\}$ and thus its expectation $\mu_{S_{-i}}(x_i)$ is in $[-1,1]$.
Intuitively, the subtraction $N(\widetilde S^{-1})-N(\widetilde S^{+1})$ implies that the prediction favors the label that leads to a smaller empirical optimum on the corrupted sequence.
Moreover, this subtraction is chosen to relate to the derivative of the expected empirical optimum of the corrupted full sequence with respect to the corruption rate $\beta$, as shown later.

We now express the leave-one-out loss using the real-valued predictions.
Let $\widetilde S^{+y_i}$ be the corrupted full sequence obtained by adding $(x_i,y_i)$ to $\widetilde S_{-i}$ at coordinate $i$, similarly define $\widetilde S^{-y_i}$ by adding $(x_i,-y_i)$ to $\widetilde S_{-i}$.
It holds that
\begin{align}
    \label{eq:loo-one-term-loss}
    \ell(\mu_{S_{-i}}(x_i),y_i)
    =\frac{1-y_i\mu_{S_{-i}}(x_i)}{2}
    =\frac{1}{2}-\frac{1}{2}\E_{\beta,\widetilde S_{-i}}[N(\widetilde S^{-y_i})-N(\widetilde S^{+y_i})].
\end{align}
With this in place, we can now prove \cref{lem:leave-one-out}.
\begin{proof}[Proof of \cref{lem:leave-one-out}]
    For every corruption rate $\beta\in[0,1/2]$, let $\widetilde S$ be the corrupted sequence obtained by flipping the labels of $S$ independently with probability $\beta$, and define the function
    \[
        \Phi(\beta)=\E_{\widetilde S}[N(\widetilde S)|\beta],
    \]
    where the conditional expectation is taken over the corrupted sequence $\widetilde S$ given the corruption rate $\beta$.
    When $\beta=0$, we have $\Phi(0)=N(S)=n\cdot\erhatmin_S(\cH)$; when $\beta=1/2$, each label is flipped independently with probability $1/2$, and thus the corrupted labels are independent uniform signs.
    Let $\bm\sigma=(\sigma_1,\ldots,\sigma_n)$ be a vector of i.i.d. Rademacher random variables, then we have
    \begin{align*}
        \Phi(1/2)
        =\E_{\bm\sigma}\left[\min_{h\in\cH_{\bm x}}\sum_{i=1}^n\ind_{\{h(x_i)\ne\sigma_i\}}\right]
        =\frac{n-\E_{\bm\sigma}\left[\max_{h\in\cH_{\bm x}}\sum_{i=1}^n\sigma_i h(x_i)\right]}{2}
        =\frac{n}{2}(1-\gRad(\cH_{\bm x})).
    \end{align*}
    We now relate the leave-one-out loss to the function $\Phi$ by differentiating $\Phi$ with respect to $\beta$.
    To differentiate $\Phi$, let $\bm\gamma=(\gamma_1,\ldots,\gamma_n)\in[0,1/2]^n$ be a vector of corruption rates, and let $\bm\gamma_{-i}$ be the vector obtained by removing the $i$-th element.
    Temporarily, suppose $\widetilde S$ is the corrupted sequence obtained by flipping the label $y_i$ of $S$ independently with probability $\gamma_i$, similarly define $\widetilde S_{-i}$, $\widetilde S^{+y_i}$ and $\widetilde S^{-y_i}$.
    Define the function
    \begin{align*}
        \Psi(\bm\gamma)
        =\E_{\widetilde S}[N(\widetilde S)|\bm\gamma].
    \end{align*}
    where the conditional expectation is taken over the corrupted sequence $\widetilde S$ given the corruption rates $\bm\gamma$.
    Then, for every $i\in[n]$, considering the two cases where $y_i$ is flipped or not flipped, we have
    \[
        \Psi(\bm\gamma)
        =\gamma_i \E_{\widetilde S_{-i}}[N(\widetilde S^{-y_i})|\bm\gamma_{-i}]+(1-\gamma_i) \E_{\widetilde S_{-i}}[N(\widetilde S^{+y_i})|\bm\gamma_{-i}],
    \]
    which implies that for every $i\in[n]$, the partial derivative of $\Psi$ with respect to $\gamma_i$ is
    \[
        \frac{\partial\Psi}{\partial \gamma_i}
        =\E_{\widetilde S_{-i}}[N(\widetilde S^{-y_i})-N(\widetilde S^{+y_i})|\bm\gamma_{-i}].
    \]
    Setting $\gamma_i=\beta$ for all $i\in[n]$, we have $\Phi(\beta)=\Psi(\gamma(\beta))$ and the multivariable chain rule gives that
    \[
        \Phi'(\beta)=\sum_{i=1}^n\frac{\partial\Psi}{\partial \gamma_i}(\beta,\ldots,\beta)\cdot \gamma_i'(\beta)
        =\sum_{i=1}^n\E_{\widetilde S_{-i}}[N(\widetilde S^{-y_i})-N(\widetilde S^{+y_i})|\beta],
    \]
    where the last equality follows from that $\gamma_i'(\beta)=1$ for all $i\in[n]$ and that every element of $\bm\gamma_{-i}$ is $\beta$.
    Writing the leave-one-out loss using \cref{eq:loo-one-term-loss}, we have
    \begin{align*}
        \frac{1}{n}\sum_{i=1}^n\ell(\mu_{S_{-i}}(x_i),y_i)
        &=\frac{1}{n}\sum_{i=1}^n\left(\frac{1}{2}-\frac{1}{2}\E_{\beta,\widetilde S_{-i}}[N(\widetilde S^{-y_i})-N(\widetilde S^{+y_i})]\right)\\
        &=\frac{1}{2}-\frac{1}{2n}\int_{0}^{1/2}2\Phi'(\beta)d\beta\\
        &=\frac{1}{2}-\frac{\Phi(1/2)-\Phi(0)}{n}\\
        &=\erhatmin_S(\cH)+\frac{\gRad(\cH_{\bm x})}{2}\\
        &\leq \erhatmin_S(\cH)+C\sqrt{\frac{d}{n}},
    \end{align*}
    where $C>0$ is a universal constant, the second last equality follows from $\Phi(0)=n\cdot\erhatmin_S(\cH)$ and $\Phi(1/2)=\frac{n}{2}(1-\gRad(\cH_{\bm x}))$, and the last inequality follows from \cref{thm:grad}.
    This concludes the proof.
\end{proof}

The prediction $\mu_{S_{-i}}(x_i)=\E_{\beta,\widetilde S_{-i}}[\nu_{\widetilde S_{-i}}(x_i)]$ can be computed exactly by enumerating all $2^{n-1}$ possible corrupted sequences $\widetilde S_{-i}$.
Fix a corrupted sequence $\widetilde S_{-i}$ obtained by flipping $k$ labels of $S_{-i}$, then the appearance probability of $\widetilde S_{-i}$ is
\[
    w_k=\int_0^{1/2}2\beta^k(1-\beta)^{n-1-k}d\beta=\frac{\sum_{j=k+1}^{n}\binom{n}{j}}{n2^{n-1}\binom{n-1}{k}}.
\]
Thus, we don't need to draw the corruption rate $\beta$, and the prediction can be computed directly as shown in \cref{alg:leave-one-out}.
A special case is that when $n=1$, $k$ must be $0$, and $w_0=1$, and there is only one possible corrupted sequence $\widetilde S_{-1}=\emptyset$.
Note that the learner doesn't depend on the coordinate information of the input sequence, and thus is permutation invariant.
\begin{algorithm}[htbp]
    \caption{Leave-one-out learner}
    \label{alg:leave-one-out}

    \KwIn{
        A held-out index $i\in[n]$,
        a training sequence
        $S_{-i}=((x_j,y_j))_{j\in[n]\setminus\{i\}}$,
        a query $x_i$, 
        and the partial concept class $\mathcal H$.
    }
    \KwOut{A prediction $\mu_{S_{-i}}(x_i)\in[-1,1]$.}

    $\mu_{S_{-i}}(x_i)\gets 0$\;
    \For{$k=0,\ldots,n-1$}{
        $w_k \gets \frac{\sum_{j=k+1}^{n}\binom{n}{j}}{n2^{n-1}\binom{n-1}{k}}$\;
        \ForEach{
            possible corrupted sequence $\widetilde S_{-i}$ with $k$ flipped labels 
        }{
            Construct $\widetilde S^{+1}, \widetilde S^{-1}$ by adding $(x_i,+1)$ and $(x_i,-1)$ to $\widetilde S_{-i}$ at coordinate $i$\; 
            Compute $N(\widetilde S^{+1})$ and $N(\widetilde S^{-1})$
            \tcp*{{\normalfont\footnotesize $N(T)=\min_{h\in\cH}\sum_{(x,y)\in T} \ind_{\{h(x)\ne y\}}$ for a labeled sequence $T$ }}
            $\nu_{\widetilde S_{-i}}(x_i)\gets N(\widetilde S^{-1})-N(\widetilde S^{+1})$\;
            $\mu_{S_{-i}}(x_i)\gets\mu_{S_{-i}}(x_i)+w_k\nu_{\widetilde S_{-i}}(x_i)$\;
        }
    }

    \Return{$\mu_{S_{-i}}(x_i)$}.
\end{algorithm}

\subsection{From leave-one-out to PAC bound}
We now adapt the aggregation technique of \cite{mathiasen2026optimal} to convert the leave-one-out bound into a PAC bound.
The idea is to run the leave-one-out learner on many subsequences of the full sequence, and then aggregate the predictions to construct a PAC learner.
\cite{DBLP:conf/focs/Aden-AliCSZ23} first used this approach to establish an optimal realizable PAC learner for total concept classes.
\cite{DBLP:conf/alt/DughmiKY25} extended it to the agnostic setting, however, obtaining a suboptimal sample complexity.
Recently, \cite{mathiasen2026optimal} improved the analysis and obtained an optimal agnostic PAC learner.

For a labeled sequence $S=((x_i,y_i))_{i=1}^n$, denote the subsequence of $S$ with first $t$ examples as $S_{\leq t}=((x_i,y_i))_{i=1}^t$, and let $S_{\leq 0}=\emptyset$.
For every $x\in\cX$, denote the prediction of the leave-one-out learner, \cref{alg:leave-one-out}, on $x$ with respect to the subsequence $S_{\leq t}$ as $\mu_{S_{\leq t}}(x)$.
We also define the average prediction of the leave-one-out learner on $x$ with respect to the subsequences $S_{\leq 0},\ldots,S_{\leq n-1}$ as
\[
    \bar \mu(x)=\frac{1}{n}\sum_{t=0}^{n-1}\mu_{S_{\leq t}}(x).
\]
Obviously, $\bar\mu(x)\in [-1,1]$.
To convert the real-valued prediction $\bar\mu(x)$ into a binary predictor, we can use a threshold $\hat\tau \in [-1,1]\cup\{-2\}$ and define the predictor as
\[
    \hat h(x)=2\ind_{\{\bar\mu(x)>\hat\tau\}}-1,
\]
where the threshold $\hat\tau$ can be selected using an independent validation set with a fixed tie-breaking rule, and the possible value $-2$ is used to cover the case where the prediction is always $1$.

The algorithm is summarized in \cref{alg:pac}, and the following theorem shows that it is an agnostic PAC learner with optimal sample complexity.
Without loss of generality, we assume that the size of the training sequence is $2n$ in \cref{alg:pac}, which only causes a constant factor in analysis.

\begin{algorithm}[htbp]
    \caption{An optimal agnostic PAC learner}
    \label{alg:pac}

    \KwIn{
        A labeled sequence $S=((x_i,y_i))_{i=1}^{2n}$ i.i.d. from the distribution $\cD$ over $\cX\times\{-1,1\}$,
        and the partial concept class $\mathcal H$.
    }
    \KwOut{A predictor $\hat{h}:\cX\to\{-1,1\}$.}
    \For{
        $t=0,\ldots,n-1$
    }{
        Let $S_{\leq t}$ be the subsequence of $S$ with the first $t$ examples
        \tcp*{{\normalfont\footnotesize $S_{\leq 0}=\emptyset$}}
        Define $\mu_{S_{\leq t}}$ as the output of the leave-one-out learner, \cref{alg:leave-one-out}, with respect to $S_{\leq t}$\;
    }
    Define the average prediction $\bar \mu=\frac{1}{n}\sum_{t=0}^{n-1}\mu_{S_{\leq t}}$\;
    \tcp{{\normalfont\footnotesize Use validation data to select a threshold}}
    Define the threshold set $\cT\gets\{\bar\mu(x_i): n<i\leq 2n\}\cup\{-2\}$\;
    $\hat\tau\gets \argmin_{\tau\in\cT}\sum_{i=n+1}^{2n} \ind_{\{2\ind_{\{\bar\mu(x_i)>\tau\}}-1\neq y_i\}}$\;
    \Return{$\hat h(x)=2\ind_{\{\bar\mu(x)>\hat\tau\}}-1$}.
\end{algorithm}

\begin{theorem}
    \label{thm:pac}
    Let $\cH\subseteq\{-1,+1,\star\}^\cX$ be a partial concept class with VC dimension $d\geq 1$, let $\cD$ be a distribution over $\cX\times\{-1,1\}$, and let $S=((x_i,y_i))_{i=1}^{n}$ be a labeled sequence i.i.d. from $\cD$.
    There exists a deterministic agnostic PAC learner, \cref{alg:pac}, which, for every $\eps,\delta\in(0,1)$, with probability at least $1-\delta$, the output predictor $\hat h$ satisfies
    \[
        \er_\cD(\hat h)\leq \ermin_\cD(\cH)+C\sqrt{\frac{d+\log(1/\delta)}{n}},
    \]
    where $C>0$ is a universal constant and $\ermin_\cD(\cH)=\min_{h\in\cH}\er_\cD(h)$.
    Thus, the sample complexity of this algorithm is $O(\frac{d+\log(1/\delta)}{\eps^2})$.
\end{theorem}
\cref{thm:pac} does not consider the case $d=0$, which is trivial since for every $x\in\cX$, all hypotheses in $\cH$ can only output $\star$ or the same label, and thus a predictor can be constructed to always output that label, which has zero excess risk.

We first use the forward and backward martingale technique of \cite{DBLP:conf/focs/Aden-AliCSZ23} to prove the following two lemmas, proofs of which are provided in \cref{sec:proof-forward,sec:proof-backward}.
We define the expected loss of $\bar\mu$ with respect to the distribution $\cD$ as $\er_\cD(\bar\mu)=\mathbb{E}_{(x,y)\sim\cD}[\ell(\bar\mu(x),y)]$.
Applying these two lemmas with the specific hypothesis in \cref{lem:backward-martingale} to be $h^*=\argmin_{h\in\cH}\er_\cD(h)$, and combining with Hoeffding's inequality to bound $|\er_\cD(h^*)-\erhat_{S}(h^*)|$, we can show that the expected loss of the average prediction $\bar\mu$ is comparable to that of the best hypothesis in $\cH$.

\begin{lemma}[Forward martingale]
    \label{lem:forward-martingale}
    Let $S=((x_i,y_i))_{i=1}^{n}$ be a labeled sequence i.i.d. from $\cD$.
    For every $\delta\in(0,1)$, with probability at least $1-\delta$, it holds that
    \[
        \er_\cD(\bar \mu)
        \leq
        \frac{1}{n}\sum_{i=0}^{n-1}\ell(\mu_{S_{\leq i}}(x_{i+1}),y_{i+1})+\sqrt{\frac{\log (1/\delta)}{2n}}.
    \]
\end{lemma}

\begin{lemma}[Backward martingale]
    \label{lem:backward-martingale}
    Let $S=((x_i,y_i))_{i=1}^{n}$ be a labeled sequence i.i.d. from $\cD$.
    For every $h\in \cH$, for every $\delta\in(0,1)$, with probability at least $1-\delta$, it holds that
    \[
        \frac{1}{n}\sum_{i=0}^{n-1}\ell(\mu_{S_{\leq i}}(x_{i+1}),y_{i+1})
        \leq
        \erhat_{S}(h)+C\sqrt{\frac{d+\log(1/\delta)}{n}},
    \]
    where $C>0$ is a universal constant and $\erhat_{S}(h)=\frac{1}{n}\sum_{i=1}^n\ind_{\{h(x_i)\neq y_i\}}$ is the empirical risk.
\end{lemma}

What remains is to show that the thresholding step doesn't increase the expected loss too much.
To this end, we provide a uniform convergence bound for the thresholded predictors, proof of which is provided in \cref{sec:proof-threshold}.

\begin{lemma}
    \label{lem:threshold}
    Let $\bar\mu:\cX\to[-1,1]$ be a real-valued function, and let $h_\tau(x)=2\ind_{\{\bar\mu(x)>\tau\}}-1$ be a predictor with respect to a threshold $\tau\in[-1,1]\cup\{-2\}$.
    Let $S=((x_i,y_i))_{i=1}^{n}$ be a labeled sequence i.i.d. from $\cD$ independent of $\bar\mu$ as a validation set.
    Then for every $\delta\in(0,1)$, with probability at least $1-\delta$, it holds that
    \[
        \sup_{\tau\in[-1,1]\cup\{-2\}}|\hat\er_{S}(h_\tau)-\er_\cD(h_\tau)|
        \leq
        7\sqrt{\frac{\log(e/\delta)}{n}}.
    \]
\end{lemma}

With these lemmas in place, we can now prove \cref{thm:pac}.
It holds that there always exists a threshold $\tau^*$ such that the expected loss of the thresholded predictor $h_{\tau^*}$ is at most that of the average prediction $\bar\mu$.
The algorithm selects a threshold $\hat\tau$ that minimizes the empirical risk on the validation set, implying the empirical risk of the thresholded predictor $h_{\hat\tau}$ is at most that of $h_{\tau^*}$.
Finally, the uniform convergence bound in \cref{lem:threshold} guarantees that the differences between these empirical risks and population risks are small.

\begin{proof}[Proof of \cref{thm:pac}]
    \cref{alg:pac} uses a labeled sequence $S$ of size $2n$, let $S_1$ be the first $n$ examples used to train these leave-one-out learners, and let $S_2$ be the last $n$ examples used as a validation set to select the threshold.
    
    On the subsequence $S_1$, applying \cref{lem:forward-martingale,lem:backward-martingale}, each with failure probability $\delta/4$, and set the specific hypothesis in \cref{lem:backward-martingale} to be $h^*=\argmin_{h\in\cH}\er_\cD(h)$, we have that with probability at least $1-\delta/2$,
    \begin{align}
        \label{eq:barmu-empirical}
        \er_\cD(\bar\mu)
        \leq
        \hat\er_{S_1}(h^*)+C_1\sqrt{\frac{d+\log(1/\delta)}{n}},
    \end{align}
    where $C_1>0$ is a universal constant.
    Applying Hoeffding's inequality to bound $|\hat\er_{S_1}(h^*)-\er_\cD(h^*)|$ with failure probability $\delta/4$, combining with \cref{eq:barmu-empirical} and using the union bound, we have that with probability at least $1-3\delta/4$,
    \begin{align}
        \label{eq:barmu}
        \er_\cD(\bar\mu)
        \leq
        \er_\cD(h^*)+C_2\sqrt{\frac{d+\log(1/\delta)}{n}}
        =
        \ermin_\cD(\cH)+C_2\sqrt{\frac{d+\log(1/\delta)}{n}},
    \end{align}
    where $C_2>0$ is a universal constant.
    Let $h_\tau(x)=2\ind_{\{\bar\mu(x)>\tau\}}-1$ be a predictor with respect to a threshold $\tau\in[-1,1]$, then for every $(x,y)\sim \cD$, we have
    \begin{align*}
        \int_{-1}^{1}\ind_{\{h_\tau(x)\neq y\}}d\tau
        &=(1-\bar\mu(x))\cdot \ind_{\{y=+1\}}+(\bar\mu(x)+1)\cdot \ind_{\{y=-1\}}\\
        &=1-y\bar\mu(x)\\
        &=2\ell(\bar\mu(x),y).
    \end{align*}
    Taking expectation over $(x,y)\sim\cD$, dividing both sides by $2$, and applying Fubini's theorem, we have
    \[
        \frac{1}{2}\E_{(x,y)\sim\cD}\left[\int_{-1}^{1}\ind_{\{h_\tau(x)\neq y\}}d\tau\right]
        =\frac{1}{2}\int_{-1}^{1}\er_\cD(h_\tau)d\tau
        =\er_\cD(\bar\mu),
    \]
    which implies that there exists a $\tau^*$ satisfying $\er_\cD(h_{\tau^*})\leq\er_\cD(\bar\mu)$.
    The threshold $\hat\tau$ is selected as the minimizer of the empirical risk on validation data $S_2$, and thus the output predictor $\hat h=h_{\hat\tau}$ satisfies
    \[
        \hat\er_{S_2}(\hat h)\leq \hat\er_{S_2}(h_{\tau^*}).
    \]
    Now applying \cref{lem:threshold} with failure probability $\delta/4$, we have that with probability at least $1-\delta/4$,
    \begin{align}
        \label{eq:threshold}
        \er_\cD(\hat h)
        &\leq \hat\er_{S_2}(\hat h)+7\sqrt{\frac{\log(4e/\delta)}{n}}\notag\\
        &\leq \hat\er_{S_2}(h_{\tau^*})+7\sqrt{\frac{\log(4e/\delta)}{n}}\notag\\
        &\leq \er_\cD(h_{\tau^*})+14\sqrt{\frac{\log(4e/\delta)}{n}}\notag\\
        &\leq \er_\cD(\bar\mu)+14\sqrt{\frac{\log(4e/\delta)}{n}}.
    \end{align}
    Combining \cref{eq:barmu,eq:threshold}, we have that with probability at least $1-\delta$,
    \begin{align*}
        \er_\cD(\hat h)
        &\leq
        \ermin_\cD(\cH)+C_2\sqrt{\frac{d+\log(1/\delta)}{n}}+14\sqrt{\frac{\log(4e/\delta)}{n}}\\
        &\leq 
        \ermin_\cD(\cH)+C_3\sqrt{\frac{d+\log(1/\delta)}{n}},
    \end{align*}
    where $C_3>0$ is a universal constant.
    The sample complexity of the algorithm directly follows from setting the excess risk to be at most $\eps$ and solving for $n$.
    Setting $n'=2n$ and scaling the parameters accordingly concludes the proof.
\end{proof}

%% file: content/a.missing_proof_grad.tex
\section{Proofs for the generalized Rademacher complexity bound}

\subsection{Verifying the approximation properties}
\label{sec:proof-approximate-vector}
\begin{lemma}[\cref{lem:grad-approximate-vector} restated]
    Let $\cH\subseteq\{-1,+1,\star\}^n$ be a partial concept class with VC dimension $d\geq 1$, $J=\lfloor \log_2n\rfloor+1$.
    For every Rademacher vector $\bm\sigma$, let $h_{\bm\sigma}$ be the maximizer of $\sum_{i\in[n]}\sigma_i h_{\bm\sigma,i}$, there exists a sequence of approximation vectors $\{\bg^j_{\bm\sigma}\}_{j=0}^J$ with $\bg^j_{\bm\sigma}\in [-1,1]^n$ for $0\leq j\leq J$, such that 
    \begin{itemize}
        \item $\bg^0_{\bm\sigma}$ is a constant vector for all $\bm\sigma$;
        \item $\norm{\bg_{\bm\sigma}^j-\bg_{\bm\sigma}^{j-1}}_2^2\leq 256n\cdot 2^{-j}$ for all $\bm\sigma$ and all $1\leq j\leq J$;
        \item $\E_{\bm\sigma}\left[\sum_{i\in[n]} \rho(g_{\bm\sigma,i}^J, h_{\bm\sigma,i})\right]\leq 0$;
        \item $|I^j|\leq \exp(Cdj^3)$ for all $1\leq j\leq J$, where $C>0$ is a universal constant. 
    \end{itemize}
\end{lemma}

\begin{proof}[Proof of \cref{lem:grad-approximate-vector}]
    \textbf{Property 1: $\bg^0_{\bm\sigma}$ is a constant vector for all $\bm\sigma$.}
    $\bg^0_{\bm\sigma}$ depends only on $p^0, q^0$ and $v^0$, which are determined by the fixed information of the disambiguations $\widebar\cH^1,\ldots, \widebar\cH^J$ and thus are the same for all $\bm\sigma$.

    \paragraph{Property 2: $\norm{\bg_{\bm\sigma}^j-\bg_{\bm\sigma}^{j-1}}_2^2\leq 256n\cdot 2^{-j}$ for all $\bm\sigma$ and all $1\leq j\leq J$.}
    When $j=1$, we have $\norm{\bg_{\bm\sigma}^1-\bg_{\bm\sigma}^0}_2^2\leq 4n\leq 256n\cdot 2^{-1}$ directly.
    
    Assume $j\geq 2$ from now on and fix $\bm\sigma$.
    We want to connect the scale $2^{-j}$ with the parameters $p^{j-1}_i,q^{j-1}_i$ and $p^j_i,q^j_i$, which determine $g^j_i, g^{j-1}_i$ and the increment.
    Let $\Pi^{j-1}$ be the uniform distribution over $\Sigma^{j-1}$, and let $\Pi^j$ be the uniform distribution over $\Sigma^j$, where $\Sigma^{j-1}$ and $\Sigma^j$ are the sets of all Rademacher vectors with the same prefix of disambiguations with the fixed $\bm\sigma$ up to step $j-1$ and $j$, respectively.

    Then $p^{j-1}_i$ equals the probability that drawing a Rademacher vector $\bm\sigma$ from $\Pi^{j-1}$, the maximizer $h_{\bm\sigma}$ has $h_{\bm\sigma,i}=+1$, and similarly for $q^{j-1}, p^j, q^j$.
    Draw $\bm \pi^{j-1}\sim \Pi^{j-1}$ and $\bm \pi^j\sim \Pi^j$ independently, then the probability that their maximizers $h^{j-1}$ and $h^j$ are both defined and have opposite values on coordinate $i$ is
    \[
        \Prob_{\bm \pi^{j-1}, \bm \pi^j}[h^{j-1}_i, h^{j}_i\neq \star \text{ and } h^{j-1}_i\neq h^j_i]
        = p^{j-1}_i q^j_i+q^{j-1}_i p^j_i.
    \]
    In terms of normalized conflict, the sum of probability over all coordinates can also be written as
    \[
        \sum_{i\in [n]}\Prob_{\bm \pi^{j-1}, \bm \pi^j}[h^{j-1}_i, h^j_i\neq \star \text{ and } h^{j-1}_i\neq h^j_i]
        =n\cdot \E_{\bm \pi^{j-1}, \bm \pi^j}[c(h^{j-1},h^j)].
    \]
    Since the two Rademacher vectors $\bm \pi^{j-1}$ and $\bm \pi^j$ have the same prefix of disambiguations $(\widebar h^1,\ldots, \widebar h^{j-1})$, their maximizers $h^{j-1}$ and $h^{j}$ have the same $2^{-j+1}$-disambiguation $\widebar h^{j-1}$, which implies that
    \[
        c(h^{j-1},h^j)
        \leq
        c(h^{j-1},\widebar h^{j-1})+c(h^j,\widebar h^{j-1})
        \leq 2\cdot 2^{-j+1}.
    \]
    Combining the above three relations, we have $\sum_{i\in [n]}(p^{j-1}_i q^j_i+q^{j-1}_i p^j_i)\leq 4n\cdot 2^{-j}$. Thus, if we can show that $|g^j_i-g^{j-1}_i|^2\leq 64(p_i^{j-1}q_i^j+q^{j-1}_i p^j_i)$ for each $i\in [n]$, the proof is complete.

    We now show that $|g^j-g^{j-1}|^2\leq 64(p^{j-1}q^j+q^{j-1}p^j)$ for a fixed coordinate $i$. Ignore the subscript $i$ for simplicity.
    \begin{itemize}
        \item If the construction doesn't stop before step $j+1$, then $v^{j-1}>1/4$ and $v^j>1/4$, thus $g^{j-1}=\theta^{j-1}$ and $g^j=\theta^j$, and we have
            \[
                |g^j-g^{j-1}|^2=(\theta^j-\theta^{j-1})^2\leq 4v^j(\theta^j-\theta^{j-1})^2;
            \]
        \item If the construction stops at step $j$, then $v^{j-1}>1/4$ and $v^j\leq 1/4$, thus $g^{j-1}=\theta^{j-1}$ and $g^j=(1-2v^j)\theta^{j-1}+2v^j\theta^j$, and we have
            \[
                |g^j-g^{j-1}|^2=|2v^j(\theta^j-\theta^{j-1})|^2= 4(v^j)^2(\theta^j-\theta^{j-1})^2\leq 4v^j(\theta^j-\theta^{j-1})^2;
            \]
        \item If the construction stops before step $j$, then $g^j=g^{j-1}$ and $|g^j-g^{j-1}|^2=0$.
    \end{itemize}
    We thus only need to consider the first two cases, where $v^{j-1}>1/4$, in which we have $|g^j-g^{j-1}|^2\leq 4v^j(\theta^j-\theta^{j-1})^2$.
    The case $v^j=0$ is trivial, and assume $v^j>0$ from now on.
    The fact that $(x-y)^2\leq 2(1-xy)$ for all $x,y\in [-1,1]$ and $1-\theta^{j-1}\theta^j=\frac{2}{v^{j-1}v^j}(p^{j-1}q^j+q^{j-1}p^j)$ gives us that
    \[
        |g^j-g^{j-1}|^2
        \leq 4v^j(\theta^j-\theta^{j-1})^2
        \leq 8v^j(1-\theta^{j-1}\theta^j)
        \leq \frac{16}{v^{j-1}}(p^{j-1}q^j+q^{j-1}p^j)
        \leq 64(p^{j-1}q^j+q^{j-1}p^j),
    \]
    where the last inequality comes from $v^{j-1}>1/4$.

    \paragraph{Property 3: $\E_{\bm\sigma}\left[\sum_{i\in[n]} \rho(g_{\bm\sigma,i}^J, h_{\bm\sigma,i})\right]\leq 0$.}
    Similarly, it suffices to show that for any fixed coordinate $i$, $\E_{\bm\sigma}[\rho(g_{\bm\sigma,i}^J, h_{\bm\sigma,i})]\leq 0$.
    Fix a coordinate $i$, let $Y$ be the random variable used to guess the value of $h_{\bm\sigma,i}$ as before.
    We partition the set of all Rademacher vectors $\{\pm 1\}^n$ into finite subsets according to the observed information $\cF^t$, where $0\leq t\leq J$ is the first step at which the construction stops.
    If the construction does not stop at any step $0,\ldots,J$, then all $g^j$ are constructed as $\theta^j$, the construction then stops at step $J+1$.

    \textbf{Case 1.} Let $\bF^J$ be a realization of the information $\cF^J$, whose corresponding construction of the sequence $\{g^j\}_{j=0}^J$ doesn't stop at any step $0,\ldots,J$, and let $\bm\Sigma^J$ be the set of all Rademacher vectors with information $\bF^J$.
    This implies that $g^j_{\bm\sigma,i}=\theta^j$ for all $0\leq j\leq J$. 
    Since $p^J q^J=0$ and $v^J>1/4$, there cannot be two Rademacher vectors in $\bm\Sigma^J$ that have opposite defined values for $h_{\bm\sigma,i}$. 
    If $p^J>0$, then $q^J=0$, $g_{\bm\sigma,i}^J=\theta^J=+1$ and $h_{\bm\sigma,i}\in\{+1,\star\}$ for all $\bm\sigma\in\bm\Sigma^J$, 
    \begin{align*}
        &\E_{\bm\sigma\in\bm\Sigma^J}[\rho(g_{\bm\sigma,i}^J, h_{\bm\sigma,i})]\\
        &=|g_{\bm\sigma,i}^J-1|\cdot \Prob_{\bm\sigma\in\bm\Sigma^J}[h_{\bm\sigma,i}=+1]
        +|g_{\bm\sigma,i}^J+1|\cdot \Prob_{\bm\sigma\in\bm\Sigma^J}[h_{\bm\sigma,i}=-1]\\
        &\quad+(|g_{\bm\sigma,i}^J|-1)\cdot \Prob_{\bm\sigma\in\bm\Sigma^J}[h_{\bm\sigma,i}=\star]\\
        &=0\cdot p^J+2\cdot q^J+(1-1)\cdot (1-v^J)\\
        &=0.
    \end{align*}
    The case $q^J>0$ is similar.

    \textbf{Case 2.} For every $1\leq t\leq J$, let $\bF^t$ be a realization of the information $\cF^t$, whose corresponding construction of the sequence stops at step $t$, and let $\bm\Sigma^t$ be the set of all Rademacher vectors with information $\bF^t$.
    This implies that $v^t\leq 1/4$ and 
    \[
        g^J_{\bm\sigma,i}
        =g^t_{\bm\sigma,i}
        =(1-2v^t)g_{\bm\sigma,i}^{t-1}+2v^t\theta^t.
    \] 
    The expected approximation error over $\bm\Sigma^t$ is
    \begin{align*}
        &\E_{\bm\sigma\in\bm\Sigma^t}[\rho(g_{\bm\sigma,i}^J, h_{\bm\sigma,i})]\\
        &=|g_{\bm\sigma,i}^t-1|\cdot \Prob_{\bm\sigma\in\bm\Sigma^t}[h_{\bm\sigma,i}=+1]
        +|g_{\bm\sigma,i}^t+1|\cdot \Prob_{\bm\sigma\in\bm\Sigma^t}[h_{\bm\sigma,i}=-1]\\
        &\quad+(|g_{\bm\sigma,i}^t|-1)\cdot \Prob_{\bm\sigma\in\bm\Sigma^t}[h_{\bm\sigma,i}=\star]\\
        &=p^t(1-g_{\bm\sigma,i}^t)+q^t(1+g_{\bm\sigma,i}^t)+(1-v^t)(|g_{\bm\sigma,i}^t|-1).
    \end{align*}
    For convenience, define the function $\phi(g_{\bm\sigma,i}^t)=p^t(1-g_{\bm\sigma,i}^t)+q^t(1+g_{\bm\sigma,i}^t)+(1-v^t)(|g_{\bm\sigma,i}^t|-1)$, which is convex with respect to $g_{\bm\sigma,i}^t$.
    Since $g_{\bm\sigma,i}^t$ is convex with respect to $g_{\bm\sigma,i}^{t-1}$ and $g_{\bm\sigma,i}^{t-1}\in[-1,1]$, thus the endpoint values of $g_{\bm\sigma,i}^t$ are
    \[
        g_{\bm\sigma,i}^t(-1)=-1+4p^t\leq 0,\quad
        g_{\bm\sigma,i}^t(1)=1-4q^t\geq 0.
    \]
    Since $\phi(g_{\bm\sigma,i}^t)$ is convex with respect to $g_{\bm\sigma,i}^t$, and the endpoint values are
    \[
        \phi(g_{\bm\sigma,i}^t(-1))=-2p^t(1-4q^t)\leq 0,\quad
        \phi(g_{\bm\sigma,i}^t(1))=-2q^t(1-4p^t)\leq 0.
    \]
    Thus, $\phi(g_{\bm\sigma,i}^t)\leq 0$ for all possible $g_{\bm\sigma,i}^t$, which means that the expected approximation error over $\bm\Sigma^t$ is nonpositive, i.e., $\E_{\bm\sigma\in\bm\Sigma^t}[\rho(g_{\bm\sigma,i}^J, h_{\bm\sigma,i})]\leq 0$.

    \textbf{Case 3.} Let $\bF^0$ be a realization of the information $\cF^0$, whose corresponding construction of the sequence stops at step $0$, and let $\bm\Sigma^0$ be the set of all Rademacher vectors with information $\bF^0$.
    This implies that $v^0\leq 1/4$ and 
    \[
        g^J_{\bm\sigma,i}
        =g^0_{\bm\sigma,i}
        =0.
    \] 
    The expected approximation error over $\bm\Sigma^0$ is
    \begin{align*}
        &\E_{\bm\sigma\in\bm\Sigma^0}[\rho(g_{\bm\sigma,i}^J, h_{\bm\sigma,i})]\\
        &=|g_{\bm\sigma,i}^0-1|\cdot \Prob_{\bm\sigma\in\bm\Sigma^0}[h_{\bm\sigma,i}=+1]
        +|g_{\bm\sigma,i}^0+1|\cdot \Prob_{\bm\sigma\in\bm\Sigma^0}[h_{\bm\sigma,i}=-1]\\
        &\quad+(|g_{\bm\sigma,i}^0|-1)\cdot \Prob_{\bm\sigma\in\bm\Sigma^0}[h_{\bm\sigma,i}=\star]\\
        &=p^0+q^0-(1-v^0)\\
        &=2v^0-1\\
        &\leq 0.
    \end{align*}
    where the last inequality comes from $v^0\leq 1/4$.

    Combining the three cases and taking the expectation over all possible realizations of the information $\cF^j$ for $0\leq j\leq J$, we have $\E_{\bm\sigma}[\rho(g_{\bm\sigma,i}^J, h_{\bm\sigma,i})]\leq 0$.

    \paragraph{Property 4: $|I^j|\leq \exp(Cdj^3)$ for all $1\leq j\leq J$, where $C>0$ is a universal constant.}
    For any increment vector $\bg_{\bm\sigma}^j-\bg_{\bm\sigma}^{j-1}$, both $\bg_{\bm\sigma}^j$ and $\bg_{\bm\sigma}^{j-1}$ are determined by the prefix of disambiguations $(\widebar h_{\bm\sigma}^1,\ldots, \widebar h_{\bm\sigma}^j)$, thus the number of possible increment vectors is bounded by the number of possible prefixes of disambiguations.

    \cref{lem:eps-disambiguation} implies that the number of possible disambiguations at scale $2^{-k}$ is bounded by
    \[
        |\widebar \cH^k|\leq \left(2^k\right)^{C_1d\log(2^k)}=\exp(C_2dk^2),
    \] 
    where $C_1,C_2>0$ are universal constants.
    Thus the number of possible prefixes of disambiguations is bounded by
    \begin{align*}
        \prod_{k=1}^j |\widebar \cH^k|
        \leq \prod_{k=1}^j \exp(C_2dk^2)
        =\exp\left(C_2d\sum_{k=1}^j k^2\right)
        =\exp\left(C_2d\cdot \frac{j(j+1)(2j+1)}{6}\right)
        \leq \exp(Cdj^3),
    \end{align*}
    where $C>0$ is a universal constant. Thus $|I^j|\leq \exp(Cdj^3)$ for all $1\leq j\leq J$ as required.
\end{proof}

\subsection{Chaining proof of the complexity bound}
\label{sec:proof-chaining}
\begin{theorem}[\cref{thm:grad} restated]
    Let $n\geq 1$ and let $\cH\subseteq\{-1,+1,\star\}^n$ be a partial concept class with VC dimension $d\geq 0$. There is a universal constant $C>0$ such that
    \[
        \gRad(\cH)\leq C\sqrt{d/n}.
    \] 
\end{theorem}
\begin{proof}[Proof of \cref{thm:grad}]
    When $d=0$, on every coordinate $i\in [n]$, every partial concept $h\in\cH$ is either undefined or has the same value.
    Thus there is a vector $\bg=(g_1,\ldots,g_n)\in\{-1,+1\}^n$ such that for every $h\in\cH$, $g_i=h_i$ whenever $h_i$ is defined.
    It follows that 
    \[
        \gRad(\cH)
        =\frac{1}{n}\E_{\bm\sigma}\left[\max_{h\in\cH}\sum_{i\in[n]}\sigma_i h_i \right]
        \leq \frac{1}{n}\E_{\bm\sigma}\left[\ip{\bm\sigma}{\bg}\right]
        =0,
    \]
    as required.

    Assume $d\geq 1$ from now on.
    Using \cref{lem:grad-approximate-vector}, for every Rademacher vector $\bm\sigma$, let $h_{\bm\sigma}$ be the maximizer of $\sum_{i\in[n]}\sigma_i h_{\bm\sigma,i}$, we construct a sequence of approximation vectors $\{\bg^j_{\bm\sigma}\}_{j=0}^J$ with $\bg^j_{\bm\sigma}\in [-1,1]^n$ for $0\leq j\leq J$.
    By definition of the approximation error $\rho(\cdot)$, we have
    \begin{align*}
        \gRad(\cH)
        &=\frac{1}{n}\E_{\bm\sigma}\left[\max_{h\in\cH}\sum_{i\in[n]}\sigma_i h_i \right]\\
        &=\frac{1}{n}\E_{\bm\sigma}\left[\langle \bm\sigma, h_{\bm\sigma} \rangle\right]\\
        &\leq\frac{1}{n}\E_{\bm\sigma}\left[\langle \bm\sigma, \bg_{\bm\sigma}^J\rangle+\sum_{i\in[n]} \rho(g_{\bm\sigma,i}^J, h_{\bm\sigma,i})\right]\\
        &\leq\frac{1}{n}\E_{\bm\sigma}\left[\langle \bm\sigma, \bg_{\bm\sigma}^J\rangle\right],
    \end{align*}
    where the last inequality comes from $\E_{\bm\sigma}\left[\sum_{i\in[n]} \rho(g_{\bm\sigma,i}^J, h_{\bm\sigma,i})\right]\leq 0$ in \cref{lem:grad-approximate-vector}.
    Now we rewrite $\bg_{\bm\sigma}^J$ as a telescoping sum of increments:
    \begin{align}
        \label{eq:grad-chaining-telescope}
        \gRad(\cH)
        &\leq\frac{1}{n} \E_{\bm\sigma}\left[\langle \bm\sigma, \bg_{\bm\sigma}^0 \rangle+\sum_{j=1}^J \langle \bm\sigma, \bg_{\bm\sigma}^j-\bg_{\bm\sigma}^{j-1}\rangle\right]\notag\\
        & \leq \frac{1}{n}\E_{\bm\sigma}\left[\sum_{j=1}^J \langle \bm\sigma, \bg_{\bm\sigma}^j-\bg_{\bm\sigma}^{j-1}\rangle\right]\notag\\
        & = \frac{1}{n}\sum_{j=1}^J\E_{\bm\sigma}\left[ \langle \bm\sigma, \bg_{\bm\sigma}^j-\bg_{\bm\sigma}^{j-1}\rangle\right],
    \end{align}
    where the second inequality comes from the fact that $\bg^0_{\bm\sigma}$ is a constant vector for all $\bm\sigma$ in \cref{lem:grad-approximate-vector}. 
    Fix $j\in [J]$, write $\bm\Delta^j_{\bm\sigma}=\bg_{\bm\sigma}^j-\bg_{\bm\sigma}^{j-1}$ for convenience, then
    \begin{align}
        \label{eq:grad-chaining-max-increment}
        \E_{\bm\sigma}\left[ \langle \bm\sigma, \bg_{\bm\sigma}^j-\bg_{\bm\sigma}^{j-1}\rangle\right]
        =\E_{\bm\sigma}\left[ \langle \bm\sigma, \bm\Delta^j_{\bm\sigma}\rangle\right]
        \leq \E_{\bm\sigma}\left[\max_{\Delta^j\in I^j} \langle \bm\sigma, \bm\Delta^j\rangle\right],
    \end{align}
    where $I^j=\{\bg_{\bm\sigma}^j-\bg_{\bm\sigma}^{j-1}: \bm\sigma\in\{\pm 1\}^n\}$ is the set of all possible increment vectors at scale $j$ as before.
    Following the classical chaining technique, let $\lambda>0$ be a parameter to be determined later, Jensen's inequality gives
    \begin{align}
        \label{eq:grad-chaining-jensen}
        \E_{\bm\sigma}\left[\max_{\Delta^j\in I^j} \langle \bm\sigma, \bm\Delta^j\rangle\right]
        =\frac{1}{\lambda}\log\exp\left(\lambda \E_{\bm\sigma}\left[\max_{\Delta^j\in I^j} \langle \bm\sigma, \bm\Delta^j\rangle\right]\right)
        \leq \frac{1}{\lambda}\log\E_{\bm\sigma}\left[\exp\left(\lambda \max_{\Delta^j\in I^j} \langle \bm\sigma, \bm\Delta^j\rangle\right)\right].
    \end{align}  
    Bounding the maximum by the sum, we have
    \begin{align}
        \label{eq:grad-chaining-sum-bound-max}
        \E_{\bm\sigma}\left[\exp\left(\lambda \max_{\Delta^j\in I^j} \langle \bm\sigma, \bm\Delta^j\rangle\right)\right]
        \leq \sum_{\Delta^j\in I^j}\E_{\bm\sigma}\left[ \exp\left(\lambda \langle \bm\sigma, \bm\Delta^j\rangle\right)\right].
    \end{align}  
    Using the independence of Rademacher variables and the fact that $\cosh(x) = \frac{e^x + e^{-x}}{2}\leq e^{x^2/2}$, we get that, for any $\bm\Delta^j\in I^j$,
    \begin{align}
        \label{eq:grad-chaining-cosh}
        \E_{\bm\sigma}\left[ \exp\left(\lambda \langle \bm\sigma, \bm\Delta^j\rangle\right)\right] 
        &= \prod_{i\in[n]}\E_{\sigma_i}\left[ \exp\left(\lambda \sigma_i \Delta^j_i\right)\right]
        = \prod_{i\in[n]}\cosh(\lambda \Delta^j_i)\notag\\
        &\leq \prod_{i\in[n]}\exp\left(\frac{\lambda^2 (\Delta^j_i)^2}{2}\right)
        = \exp\left(\frac{\lambda^2 \norm{\bm\Delta^j}_2^2}{2}\right)
        \leq \exp\left(\frac{\lambda^2 \max_{\Delta^j\in I^j}\norm{\bm\Delta^j}_2^2}{2}\right).
    \end{align}  
    Combining \cref{eq:grad-chaining-jensen,eq:grad-chaining-sum-bound-max,eq:grad-chaining-cosh}, we have
    \[
        \E_{\bm\sigma}\left[\max_{\Delta^j\in I^j} \langle \bm\sigma, \bm\Delta^j\rangle\right]
        \leq 
        \frac{1}{\lambda}\log\left(|I^j|\cdot \exp\left(\frac{\lambda^2 \max_{\Delta^j\in I^j}\norm{\bm\Delta^j}_2^2}{2}\right)\right)
        = \frac{\log|I^j|}{\lambda}+\frac{\lambda \max_{\Delta^j\in I^j}\norm{\bm\Delta^j}_2^2}{2}.
    \]
    If $|I^j|=1$, then $\E_{\bm\sigma}\left[\max_{\Delta^j\in I^j} \langle \bm\sigma, \bm\Delta^j\rangle\right]=0$ and it doesn't contribute to the sum \cref{eq:grad-chaining-telescope}.
    Assume $|I^j|>1$, and choose $\lambda=\sqrt{2\log|I^j|/\max_{\Delta^j\in I^j}\norm{\bm\Delta^j}_2^2}$, we have
    \[
        \E_{\bm\sigma}\left[\max_{\Delta^j\in I^j} \langle \bm\sigma, \bm\Delta^j\rangle\right]
        \leq \sqrt{2\log|I^j|\cdot \max_{\Delta^j\in I^j}\norm{\bm\Delta^j}_2^2}.
    \]
    \cref{lem:grad-approximate-vector} gives us that $\max_{\Delta^j\in I^j}\norm{\bm\Delta^j}_2^2\leq 256n\cdot 2^{-j}$ and $|I^j|\leq \exp(C_1dj^3)$, where $C_1>0$ is a universal constant. Thus
    \[
        \E_{\bm\sigma}\left[\max_{\Delta^j\in I^j} \langle \bm\sigma, \bm\Delta^j\rangle\right]
        \leq \sqrt{2\cdot C_1dj^3\cdot 256n\cdot 2^{-j}}
        =C_2\sqrt{dnj^3\cdot 2^{-j}},
    \]
    where $C_2>0$ is a universal constant. 
    Plugging this bound into \cref{eq:grad-chaining-telescope,eq:grad-chaining-max-increment}, we have
    \begin{align*}
        \gRad(\cH)
        &\leq \frac{1}{n}\sum_{j=1}^J\E_{\bm\sigma}\left[ \langle \bm\sigma, \bg_{\bm\sigma}^j-\bg_{\bm\sigma}^{j-1}\rangle\right]\\
        &\leq \frac{1}{n}\sum_{j=1}^J C_2\sqrt{dnj^3\cdot 2^{-j}}\\
        &=C_2\sqrt{\frac{d}{n}}\sum_{j=1}^J j^{3/2}\cdot 2^{-j/2}\\
        &\leq C\sqrt{\frac{d}{n}},
    \end{align*}
    where $C>0$ is a universal constant, and the last inequality comes from the fact that $\sum_{j=1}^\infty j^{3/2}\cdot 2^{-j/2}< 19$.
    This concludes the proof.
\end{proof}

%% file: content/b.missing_proof_pac.tex
\section{Proofs for the agnostic PAC learner}

\subsection{Forward martingale}
\label{sec:proof-forward}
\begin{lemma}[Forward martingale, \cref{lem:forward-martingale} restated]
    Let $S=((x_i,y_i))_{i=1}^{n}$ be a labeled sequence i.i.d. from $\cD$.
    For every $\delta\in(0,1)$, with probability at least $1-\delta$, it holds that
    \[
        \er_\cD(\bar \mu)
        \leq
        \frac{1}{n}\sum_{i=0}^{n-1}\ell(\mu_{S_{\leq i}}(x_{i+1}),y_{i+1})+\sqrt{\frac{\log (1/\delta)}{2n}}.
    \]
\end{lemma}

\begin{proof}[Proof of \cref{lem:forward-martingale}]
    Since $\bar\mu=\frac{1}{n}\sum_{i=0}^{n-1}\mu_{S_{\leq i}}$ and $\ell(a,b)=\frac{1-ab}{2}$, we have
    \begin{align*}
        \er_\cD(\bar \mu)
        =\E_{(x,y)\sim\cD}\left[\frac{1-y\cdot\frac{1}{n}\sum_{i=0}^{n-1}\mu_{S_{\leq i}}(x)}{2}\right]
        =\E_{(x,y)\sim\cD}\left[\frac{1}{n}\sum_{i=0}^{n-1}\frac{1-y\cdot\mu_{S_{\leq i}}(x)}{2}\right]
        =\frac{1}{n}\sum_{i=0}^{n-1}\er_\cD(\mu_{S_{\leq i}}).
    \end{align*}
    It thus suffices to show that with probability at least $1-\delta$, 
    \[
        \frac{1}{n}\sum_{i=0}^{n-1}\er_\cD(\mu_{S_{\leq i}})
        \leq
        \frac{1}{n}\sum_{i=0}^{n-1}\ell(\mu_{S_{\leq i}}(x_{i+1}),y_{i+1})+\sqrt{\frac{\log (1/\delta)}{2n}}.
    \]

    Suppose the sequence $S$ is generated by iteratively drawing $(x_i,y_i)$ from $\cD$ for $i=1,\ldots,n$, define a forward filtration $(\cF_i)_{i=0}^n$ such that $\cF_0$ is trivial and $\cF_i=\sigma((x_1,y_1),\ldots,(x_i,y_i))$ contains the information of the first $i$ examples $(x_j,y_j)$ for $j=1,\ldots,i$.
    
    At step $t$ with $1\leq t\leq n$, before seeing the $t$-th example $(x_t,y_t)$, namely, conditional on $\cF_{t-1}$, since $(x_t,y_t)$ is drawn from $\cD$, we have
    \[
        \E_{(x_t,y_t)\sim\cD}[\ell(\mu_{S_{\leq t-1}}(x_t),y_t)|\cF_{t-1}]
        =\er_\cD(\mu_{S_{\leq t-1}}).
    \]
    Define a random variable 
    \[
        \Delta_t=\er_\cD(\mu_{S_{\leq t-1}})-\ell(\mu_{S_{\leq t-1}}(x_t),y_t),
    \]
    which is the difference between the loss of the leave-one-out learner trained on $S_{\leq t-1}$ and its expected loss at step $t$.
    Then, $\Delta_t$ is $\cF_t$-measurable and $\E[\Delta_t|\cF_{t-1}]=0$, which implies that $(\Delta_t)_{t=1}^n$ is a martingale difference sequence with respect to the filtration $(\cF_t)_{t=0}^n$.
    Besides, since $\ell(a,b)\in[0,1]$ for any $a,b\in[-1,1]$, the range of $\Delta_t$ is $[\er_\cD(\mu_{S_{\leq t-1}})-1,\er_\cD(\mu_{S_{\leq t-1}})]$, thus the range length is $1$.
    Applying the Azuma-Hoeffding inequality (\cref{lem:azuma-hoeffding}) to the martingale difference sequence $(\Delta_t)_{t=1}^n$, we have that with probability at least $1-\delta$,
    \[
        \sum_{t=1}^n\Delta_t
        \leq \sqrt{\frac{n\log(1/\delta)}{2}}.
    \]
    Combining this with 
    \[
        \sum_{t=1}^n\Delta_t
        =\sum_{t=1}^n\er_\cD(\mu_{S_{\leq t-1}})-\sum_{t=1}^n\ell(\mu_{S_{\leq t-1}}(x_t),y_t)
        =\sum_{i=0}^{n-1}\er_\cD(\mu_{S_{\leq i}})-\sum_{i=0}^{n-1}\ell(\mu_{S_{\leq i}}(x_{i+1}),y_{i+1}),
    \]
    dividing both sides by $n$, and rearranging the terms gives the desired bound.
    This concludes the proof.
\end{proof}

\subsection{Backward martingale}
\label{sec:proof-backward}

\begin{lemma}[Backward martingale, \cref{lem:backward-martingale} restated]
    Let $S=((x_i,y_i))_{i=1}^{n}$ be a labeled sequence i.i.d. from $\cD$.
    For every $h\in \cH$, for every $\delta\in(0,1)$, with probability at least $1-\delta$, it holds that
    \[
        \frac{1}{n}\sum_{i=0}^{n-1}\ell(\mu_{S_{\leq i}}(x_{i+1}),y_{i+1})
        \leq
        \erhat_{S}(h)+C\sqrt{\frac{d+\log(1/\delta)}{n}},
    \]
    where $C>0$ is a universal constant and $\erhat_{S}(h)=\frac{1}{n}\sum_{i=1}^n\ind_{\{h(x_i)\neq y_i\}}$ is the empirical risk.
\end{lemma}

\begin{proof}[Proof of \cref{lem:backward-martingale}]
    Let $\pi$ be a uniformly random permutation of $[n]$, and let $T=((x_{\pi(i)},y_{\pi(i)}))_{i=1}^n$ be the permuted sequence.
    Since permutation does not change the distribution of the sequence, $T$ is also i.i.d. from $\cD$.
    Thus, it suffices to show that with probability at least $1-\delta$, $T$ satisfies
    \[
        \frac{1}{n}\sum_{i=0}^{n-1}\ell(\mu_{T_{\leq i}}(x_{\pi(i+1)}),y_{\pi(i+1)})
        \leq
        \erhat_{T}(h)+C\sqrt{\frac{d+\log(1/\delta)}{n}},
    \]
    where $C>0$ is a universal constant and $T_{\leq i}$ be the subsequence of $T$ with the first $i$ examples.

    We now define a filtration with respect to $T$.
    Suppose these permuted indices are revealed in reverse order, i.e., reveal $\pi(n)$ first, then $\pi(n-1)$, and so on, until $\pi(1)$ is revealed last.
    For $t=0,\ldots,n$, let $\cF_t=\sigma(S,\pi(n),\ldots,\pi(n-t+1))$ contain the information of the original sequence $S$ and the revealed indices $\pi(n),\ldots,\pi(n-t+1)$, and let $\cF_0=\sigma(S)$.

    At step $t$ with $1\leq t\leq n$, put $k=n-t+1$, then $\pi(k)$ is the revealed index at this step.
    Though the internal order of $T_{\leq k-1}$ is unknown, the leave-one-out learner $\mu_{T_{\leq k-1}}$ can be computed since it's permutation invariant.
    Define the loss difference between the loss of the leave-one-out learner trained on $T_{\leq k-1}$ and the loss of $h$ at step $t$ as 
    \[ 
        D_t = \ell(\mu_{T_{\leq k-1}}(x_{\pi(k)}),y_{\pi(k)})-\ind_{\{h(x_{\pi(k)})\neq y_{\pi(k)}\}},
    \]
    which is also a random variable. Then, we have
    \begin{align}
        \label{eq:backward-sum-loss-difference}
        \sum_{t=1}^{n}D_t
        &=\sum_{i=0}^{n-1}\ell(\mu_{T_{\leq i}}(x_{\pi(i+1)}),y_{\pi(i+1)})-\sum_{i=1}^n\ind_{\{h(x_{\pi(i)})\neq y_{\pi(i)}\}}\notag\\
        &=\sum_{i=0}^{n-1}\ell(\mu_{T_{\leq i}}(x_{\pi(i+1)}),y_{\pi(i+1)})-n\cdot\erhat_{T}(h).
    \end{align}
    Thus, we only need to prove that with probability at least $1-\delta$, $\frac{1}{n}\sum_{t=1}^{n}D_t \leq C\sqrt{\frac{d+\log(1/\delta)}{n}}$, which will be shown by considering the expectation of $D_t$ conditional on $\cF_{t-1}$ and applying the Azuma-Hoeffding inequality.

    At step $t$, before seeing the revealed index $\pi(k)$, namely, conditional on $\cF_{t-1}$, since the permutation is uniform, $(x_{\pi(k)}, y_{\pi(k)})$ is thus uniformly distributed over $T_{\leq k}$, which implies that
    \begin{align*}
        \E_{\pi}[\ind_{\{h(x_{\pi(k)})\neq y_{\pi(k)}\}}|\cF_{t-1}]
        &=\frac{1}{k}\sum_{i\in [k]}\ind_{\{h(x_{\pi(i)})\neq y_{\pi(i)}\}},\\
        \E_{\pi}[\ell(\mu_{T_{\leq k-1}}(x_{\pi(k)}),y_{\pi(k)})|\cF_{t-1}]
        &=\frac{1}{k}\sum_{i\in [k]}\ell(\mu_{T_{\leq k,-i}}(x_{\pi(i)}),y_{\pi(i)}),
    \end{align*}
    where $T_{\leq k,-i}$ is the subsequence of $T_{\leq k}$ with the $i$-th example $(x_{\pi(i)},y_{\pi(i)})$ removed.
    Combining the above two equations and applying the leave-one-out bound in \cref{lem:leave-one-out} on the subsequence $T_{\leq k}$, we have
    \begin{align}
        \label{eq:backward-expected-sum-loss-difference}
        \E_{\pi}[D_t|\cF_{t-1}]
        &=\frac{1}{k}\sum_{i\in [k]}\ell(\mu_{T_{\leq k,-i}}(x_{\pi(i)}),y_{\pi(i)})-\frac{1}{k}\sum_{i\in [k]}\ind_{\{h(x_{\pi(i)})\neq y_{\pi(i)}\}}\notag\\
        &\leq \erhatmin_{T_{\leq k}}(\cH)+C_1\sqrt{\frac{d}{k}}-\erhat_{T_{\leq k}}(h)\notag\\
        &\leq C_1\sqrt{\frac{d}{k}},
    \end{align}
    where $C_1>0$ is a universal constant and the last inequality follows from the fact that $\erhatmin_{T_{\leq k}}(\cH)\leq \erhat_{T_{\leq k}}(h)$ by the definition of minimum empirical risk.

    What remains is to bound the difference between $D_t$ and its conditional expectation $\E_{\pi}[D_t|\cF_{t-1}]$.
    Now, for $1\leq t\leq n$, define a random variable 
    \[
        \Delta_t = D_t - \E_{\pi}[D_t|\cF_{t-1}].
    \]
    Thus, $\Delta_t$ is $\cF_t$-measurable and has zero mean conditional on $\cF_{t-1}$, which means that $\Delta_1,\ldots,\Delta_n$ is a martingale difference sequence with respect to the filtration $\cF_0,\ldots,\cF_n$.
    Besides, $D_t$ is bounded in $[-1,1]$, thus the range length of $\Delta_t$ is at most $2$.
    Applying the Azuma-Hoeffding inequality in \cref{lem:azuma-hoeffding}, it holds that, with probability at least $1-\delta$,
    \begin{align}
        \label{eq:backward-azuma-hoeffding}
        \sum_{t=1}^{n}\Delta_t=\sum_{t=1}^{n}D_t-\sum_{t=1}^{n}\E_{\pi}[D_t|\cF_{t-1}]
        \leq \sqrt{2n\log(1/\delta)}.
    \end{align}
    Combining \cref{eq:backward-sum-loss-difference,eq:backward-expected-sum-loss-difference,eq:backward-azuma-hoeffding}, rearranging, and dividing both sides by $n$, we have that with probability at least $1-\delta$,
    \begin{align*}
        \frac{1}{n}\sum_{i=0}^{n-1}\ell(\mu_{T_{\leq i}}(x_{\pi(i+1)}),y_{\pi(i+1)})
        &\leq \erhat_{T}(h)+\frac{1}{n}\sum_{t=1}^{n}\E_{\pi}[D_t|\cF_{t-1}]+\sqrt{\frac{2\log(1/\delta)}{n}}\\
        &\leq \erhat_{T}(h)+\frac{C_1}{n}\sum_{k=1}^{n}\sqrt{\frac{d}{k}}+\sqrt{\frac{2\log(1/\delta)}{n}}\\
        &\leq \erhat_{T}(h)+C_2\sqrt{\frac{d+\log(1/\delta)}{n}},
    \end{align*}
    where $C_2>0$ is a universal constant and the last inequality follows from the fact that $\sum_{k=1}^{n}\sqrt{\frac{1}{k}}\leq 2\sqrt{n}$.
    This concludes the proof.
\end{proof}

\subsection{Uniform convergence for threshold predictors}
\label{sec:proof-threshold}

\begin{lemma}[\cref{lem:threshold} restated]
    Let $\bar\mu:\cX\to[-1,1]$ be a real-valued function, and let $h_\tau(x)=2\ind_{\{\bar\mu(x)>\tau\}}-1$ be a predictor with respect to a threshold $\tau\in[-1,1]\cup\{-2\}$.
    Let $S=((x_i,y_i))_{i=1}^{n}$ be a labeled sequence i.i.d. from $\cD$ independent of $\bar\mu$ as a validation set.
    Then for every $\delta\in(0,1)$, with probability at least $1-\delta$, it holds that
    \[
        \sup_{\tau\in[-1,1]\cup\{-2\}}|\hat\er_{S}(h_\tau)-\er_\cD(h_\tau)|
        \leq
        7\sqrt{\frac{\log(e/\delta)}{n}}.
    \]
\end{lemma}

\begin{proof}[Proof of \cref{lem:threshold}]
    The proof starts with a standard symmetrization argument, let $\bm\sigma=(\sigma_1,\ldots,\sigma_n)$ be a vector of i.i.d. Rademacher random variables, and let $S'=((x_i',y_i'))_{i=1}^{n}$ be an independent copy of $S$, then we have
    \begin{align*}
        \E_{S}\left[\sup_{\tau}\left|\hat\er_{S}(h_\tau)-\er_\cD(h_\tau)\right|\right]
        &=\E_{S}\left[\sup_{\tau}\left|\hat\er_{S}(h_\tau)-\E_{S'}[\hat\er_{S'}(h_\tau)]\right|\right]\\
        &\leq\E_{S,S'}\left[\sup_{\tau}\left|\hat\er_{S}(h_\tau)-\hat\er_{S'}(h_\tau)\right|\right]\\
        &=\frac{1}{n}\E_{\bm\sigma,S,S'}\left[\sup_{\tau}\left|\sum_{i=1}^n\sigma_i\left(\ind_{\{h_\tau(x_i')\neq y_i'\}}-\ind_{\{h_\tau(x_i)\neq y_i\}}\right)\right|\right]\\
        &\leq \frac{1}{n}\E_{\bm\sigma,S'}\left[\sup_{\tau}\left|\sum_{i=1}^n\sigma_i\ind_{\{h_\tau(x_i')\neq y_i'\}}\right|\right]+\frac{1}{n}\E_{\bm\sigma,S}\left[\sup_{\tau}\left|\sum_{i=1}^n\sigma_i\ind_{\{h_\tau(x_i)\neq y_i\}}\right|\right]\\
        &=\frac{2}{n}\E_{\bm\sigma,S}\left[\sup_{\tau}\left|\sum_{i=1}^n\sigma_i\ind_{\{h_\tau(x_i)\neq y_i\}}\right|\right],
    \end{align*}
    where the first inequality follows from Jensen's inequality, the second inequality follows from the triangle inequality, and the last equality follows from the fact that $S$ and $S'$ are both i.i.d. from $\cD$.
    
    Now, we bound the last term to bound $\E_{S}\left[\sup_{\tau}\left|\hat\er_{S}(h_\tau)-\er_\cD(h_\tau)\right|\right]$, then the final bound follows from McDiarmid's inequality. 
    For every fixed $S$, reorder the examples in $S$ according to the values of $\bar\mu(x_i)$ so that $\bar\mu(x_1)\geq\cdots\geq\bar\mu(x_n)$.
    For each $\tau\in[-1,1]\cup \{-2\}$, let $k(\tau)=\max\{i\in[n]:\bar\mu(x_i)>\tau\}$ and set $k(\tau)=0$ if $\bar\mu(x_i)\leq\tau$ for all $i\in[n]$, then we have $h_\tau(x_i)=+1$ for $i\leq k(\tau)$ and $h_\tau(x_i)=-1$ for $i>k(\tau)$.
    Note that 
    \[ 
        \ind_{\{h_\tau(x_i)\neq y_i\}}=\frac{1+y_i}{2}-y_i\ind_{\{\bar\mu(x_i)>\tau\}},
    \]
    and thus,
    \begin{align}
        \label{eq:val-two-terms}
        \sup_\tau\left|\sum_{i=1}^n\sigma_i\ind_{\{h_\tau(x_i)\neq y_i\}}\right|
        &=\sup_{\tau}\left|\sum_{i=1}^n\sigma_i\frac{1+y_i}{2}-\sum_{i=1}^{k(\tau)}\sigma_iy_i\right|\notag\\
        &\leq \left|\sum_{i=1}^n\sigma_i\frac{1+y_i}{2}\right|+\max_{0\leq k\leq n}\left|-\sum_{i=1}^{k}\sigma_iy_i\right|.
    \end{align}
    We only need to bound the expectation of supremum over these two terms separately.
    The first term doesn't depend on $\tau$, so we can bound it directly. 
    Cauchy-Schwarz inequality gives
    \begin{align}
        \label{eq:val-first-term}
        \E_{\bm\sigma}\left[\left|\sum_{i=1}^n\sigma_i\frac{1+y_i}{2}\right|\right]
        \leq \sqrt{\E_{\bm\sigma}\left[\left(\sum_{i=1}^n\sigma_i\frac{1+y_i}{2}\right)^2\right]}
        =\sqrt{\sum_{i=1}^n\left(\frac{1+y_i}{2}\right)^2}
        \leq \sqrt{n},
    \end{align}
    where the equality follows from the fact that these Rademacher variables are independent and have zero mean, implying that for $i\neq j$, $\E_{\bm\sigma}[\sigma_i\sigma_j\frac{1+y_i}{2}\frac{1+y_j}{2}]=0$.

    For the second term, define $\Delta_k=-\sum_{i=1}^k\sigma_iy_i$ for $k=0,1,\ldots,n$ with $\Delta_0=0$, then we need to bound $\E_{\bm\sigma}[\max_{0\leq k\leq n}|\Delta_k|]$.
    Since $S$ is fixed, $y_i\in\{-1,1\}$ and $\sigma_i$ are i.i.d. Rademacher random variables, $\Delta_k$ is a martingale with respect to the filtration $\cF_k=\sigma(\sigma_1,\ldots,\sigma_k)$ with $\cF_0$ trivial.
    Then, $\{\Delta_k^2\}_{k=0}^n$ is a nonnegative submartingale with respect to the same filtration.
    Besides, it holds that
    \[
        \E_{\bm\sigma}[\Delta^2_n]
        =\E_{\bm\sigma}\left[\sum_{i=1}^n\sigma_i^2y_i^2\right]+\E_{\bm\sigma}\left[\sum_{i\neq j}\sigma_i\sigma_j y_iy_j\right]
        =\E_{\bm\sigma}\left[\sum_{i=1}^n\sigma_i^2y_i^2\right]
        =n,
    \]
    where the second equality follows from the fact that these Rademacher variables are independent and have zero mean.
    Thus, Cauchy-Schwarz inequality and Doob's $L^2$ maximal inequality in \cref{lem:doob} give that
    \begin{align}
        \label{eq:val-second-term}
        \E_{\bm\sigma}[\max_{0\leq k\leq n}|\Delta_k|]
        \leq \sqrt{\E_{\bm\sigma}[\max_{0\leq k\leq n}\Delta_k^2]}
        \leq 2\sqrt{\E_{\bm\sigma}[\Delta_n^2]}
        =2\sqrt{n}.
    \end{align}
    Noting that \cref{eq:val-two-terms,eq:val-first-term,eq:val-second-term} hold for every fixed $S$, we have
    \[
        \E_{S}\left[\sup_{\tau}\left|\hat\er_{S}(h_\tau)-\er_\cD(h_\tau)\right|\right]
        \leq\frac{2}{n}\E_{\bm\sigma,S}\left[\sup_{\tau}\left|\sum_{i=1}^n\sigma_i\ind_{\{h_\tau(x_i)\neq y_i\}}\right|\right]
        \leq \frac{6}{\sqrt{n}}.
    \]
    Finally, since changing one example in $S$ can change the supremum by at most $1/n$, applying McDiarmid's inequality in \cref{lem:mcdiarmid} gives that with probability at least $1-\delta$,
    \[
        \sup_{\tau}|\hat\er_{S}(h_\tau)-\er_\cD(h_\tau)|
        \leq
        \E_{S}\left[\sup_{\tau}|\hat\er_{S}(h_\tau)-\er_\cD(h_\tau)|\right]+\sqrt{\frac{\log(1/\delta)}{2n}}
        \leq
        7\sqrt{\frac{\log(e/\delta)}{n}}.
    \]
    This concludes the proof.
\end{proof}

\subsection{An alternative definition of agnostic PAC learning}\label{sec:alter-agnostic}
\cite[Definition 38]{DBLP:conf/focs/AlonHHM21} defines the goal of agnostic PAC learning as finding a concept $\hat h$ such that with probability at least $1-\delta$, 
\[
    \er_\cD(\hat h)\leq \lim_{m\to\infty}\eps^\star(m)+\eps,
\] 
where $\eps^\star(m)=\E_{S\sim \cD^m}[\min_{h\in\cH}\erhat_S(h)]$ is the expected minimum empirical risk over $m$ samples.

In the following, we adapt the proof of \cref{thm:pac} to show that our \cref{alg:pac} also satisfies this definition of agnostic PAC learning.
The modification is that, we apply \cref{lem:backward-martingale} with the empirical risk minimizer $h_1\in\argmin_{h\in\cH}\erhat_{S_1}(h)$, which is a random variable depending on $S_1$, and apply McDiarmid's inequality to bound $\left|\hat\er_{S_1}(h_1)-\E_{S_1}\left[\hat\er_{S_1}(h_1)\right]\right|$.

\paragraph{Proof.} \cref{alg:pac} uses a labeled sequence $S$ of size $2n$, let $S_1$ be the first $n$ examples used to train these leave-one-out learners, and let $S_2$ be the last $n$ examples used as a validation set to select the threshold.

\cref{lem:backward-martingale} is applicable to any fixed hypothesis $h\in\cH$ over the randomness of $S_1$.
However, we want to apply it to $h_1$ that is the empirical risk minimizer of $S_1$, which depends on $S_1$.
Luckily, the proof of \cref{lem:backward-martingale} starts with a random permutation of the sequence $S_1$, and the empirical risk minimizer $h_1$ is invariant to the order of the examples in $S_1$, so we can fix it before the permutation, thus the initial $\sigma$-algebra $\cF_0$ in the proof already contains $h_1=h_1(S_1)$, and the proof still holds.

Combining with \cref{lem:forward-martingale}, each with failure probability $\delta/4$, we have that with probability at least $1-\delta/2$,
\begin{align}
    \label{eq:barmu-empirical-2}
    \er_\cD(\bar\mu)
    \leq
    \hat\er_{S_1}(h_1)+C_1\sqrt{\frac{d+\log(1/\delta)}{n}},
\end{align}
where $C_1>0$ is a universal constant.
Since $h_1$ is the empirical risk minimizer on $S_1$, we see $\erhat_{S_1}(h_1)$ as a function of $S_1$.
Changing one example in $S_1$ can change the empirical risk $\hat\er_{S_1}(h_1)$ by at most $1/n$, applying McDiarmid's inequality in \cref{lem:mcdiarmid} with failure probability $\delta/4$, we have that with probability at least $1-\delta/4$, 
\[
    \hat\er_{S_1}(h_1)\leq \E_{S_1}[\hat\er_{S_1}(h_1)]+\sqrt{\frac{\log(4/\delta)}{2n}}
    =\eps^\star(n)+\sqrt{\frac{\log(4/\delta)}{2n}}
    \leq\lim_{m\to\infty}\eps^\star(m)+\sqrt{\frac{\log(4/\delta)}{2n}}
\]
where the last inequality follows from the fact that $\eps^\star(m)$ is non-decreasing for $m\geq 2$.
Combining with \cref{eq:barmu-empirical-2} and using the union bound, we have that with probability at least $1-3\delta/4$,
\begin{align}
    \label{eq:barmu-2}
    \er_\cD(\bar\mu)
    \leq
    \lim_{m\to\infty}\eps^\star(m)+C_2\sqrt{\frac{d+\log(1/\delta)}{n}},
\end{align}
where $C_2>0$ is a universal constant.
Let $h_\tau(x)=2\ind_{\{\bar\mu(x)>\tau\}}-1$ be a predictor with respect to a threshold $\tau\in[-1,1]$, then for every $(x,y)\sim \cD$, we have
\begin{align*}
    \int_{-1}^{1}\ind_{\{h_\tau(x)\neq y\}}d\tau
    &=(1-\bar\mu(x))\cdot \ind_{\{y=+1\}}+(\bar\mu(x)+1)\cdot \ind_{\{y=-1\}}\\
    &=1-y\bar\mu(x)\\
    &=2\ell(\bar\mu(x),y).
\end{align*}
Taking expectation over $(x,y)\sim\cD$, dividing both sides by $2$, and applying Fubini's theorem, we have
\[
    \frac{1}{2}\E_{(x,y)\sim\cD}\left[\int_{-1}^{1}\ind_{\{h_\tau(x)\neq y\}}d\tau\right]
    =\frac{1}{2}\int_{-1}^{1}\er_\cD(h_\tau)d\tau
    =\er_\cD(\bar\mu),
\]
which implies that there exists a $\tau^*$ satisfying $\er_\cD(h_{\tau^*})\leq\er_\cD(\bar\mu)$.
The threshold $\hat\tau$ is selected as the minimizer of the empirical risk on validation data $S_2$, and thus the output predictor $\hat h=h_{\hat\tau}$ satisfies
\[
    \hat\er_{S_2}(\hat h)\leq \hat\er_{S_2}(h_{\tau^*}).
\]
Now applying \cref{lem:threshold} with failure probability $\delta/4$, we have that with probability at least $1-\delta/4$,
\begin{align}
    \label{eq:threshold-2}
    \er_\cD(\hat h)
    &\leq \hat\er_{S_2}(\hat h)+7\sqrt{\frac{\log(4e/\delta)}{n}}\notag\\
    &\leq \hat\er_{S_2}(h_{\tau^*})+7\sqrt{\frac{\log(4e/\delta)}{n}}\notag\\
    &\leq \er_\cD(h_{\tau^*})+14\sqrt{\frac{\log(4e/\delta)}{n}}\notag\\
    &\leq \er_\cD(\bar\mu)+14\sqrt{\frac{\log(4e/\delta)}{n}}.
\end{align}
Combining \cref{eq:barmu-2,eq:threshold-2}, we have that with probability at least $1-\delta$,
\begin{align*}
    \er_\cD(\hat h)
    &\leq
    \lim_{m\to\infty}\eps^\star(m)+C_2\sqrt{\frac{d+\log(1/\delta)}{n}}+14\sqrt{\frac{\log(4e/\delta)}{n}}\\
    &\leq 
    \lim_{m\to\infty}\eps^\star(m)+C_3\sqrt{\frac{d+\log(1/\delta)}{n}},
\end{align*}
where $C_3>0$ is a universal constant.
The sample complexity of the algorithm directly follows from setting the excess risk to be at most $\eps$ and solving for $n$.
Setting $n'=2n$ and scaling the parameters accordingly concludes the proof.

\subsection{Auxiliary inequalities}

\begin{lemma}[Azuma-Hoeffding inequality \cite{azuma1967weighted}]
    \label{lem:azuma-hoeffding}
    Let $(X_i)_{i=1}^n$ be a martingale difference sequence with respect to a filtration $(\cF_i)_{i=0}^n$. 
    Suppose that for every $i\in[n]$, $a_i \leq X_i \leq b_i$, $b_i-a_i\leq c_i$ almost surely, where $a_i,b_i\in\mathbb R$ are $\cF_{i-1}$-measurable and $c_i> 0$ is a deterministic constant.
    Then, for every $\eps>0$,
    \[
        \Prob\left[\sum_{i=1}^n X_i\geq \eps\right]\leq \exp\left(-\frac{2\eps^2}{\sum_{i=1}^n c_i^2}\right).
    \]
    The same bound holds for $\Prob\left[\sum_{i=1}^n X_i\leq -\eps\right]$.
\end{lemma}

\begin{lemma}[Doob's $L^2$ maximal inequality \cite{Doob1953}]
    \label{lem:doob}
    Let $\{X_i\}_{i=0}^n$ be a martingale with $\mathbb E[X_n^2]<\infty$. Then
    \[
        \mathbb E\left[\max_{1\leq i\leq n}|X_i|^2\right]\leq 4\mathbb E[|X_n|^2].
    \]
\end{lemma}

\begin{lemma}[McDiarmid's inequality \cite{mcdiarmid1989method}]
    \label{lem:mcdiarmid}
    Let $X_1,\ldots,X_n$ be independent random variables with $X_i$ taking values in a set $\mathcal X_i$, and let $f:\mathcal X_1\times\cdots\times\mathcal X_n\to\mathbb R$ be a function.
    Suppose that there exist constants $c_1,\ldots,c_n\geq 0$ such that, for every $i\in[n]$, if two realizations $(x_1,\ldots,x_n)$ and $(x_1,\ldots,x_i',\ldots,x_n)$ differ only in the $i$-th coordinate, then
    \[
        |f(x_1,\ldots,x_i,\ldots,x_n)-f(x_1,\ldots,x_i',\ldots,x_n)|\leq c_i.
    \]
    Then, for every $\eps>0$,
    \[
        \Prob\left[f(X_1,\ldots,X_n)-\mathbb E\left[ f(X_1,\ldots,X_n)\right]\geq \eps\right]
        \leq
        \exp\left(-\frac{2\eps^2}{\sum_{i=1}^n c_i^2}\right),
    \]
    the same bound holds for $\Prob\left[\mathbb E\left[ f(X_1,\ldots,X_n)\right]-f(X_1,\ldots,X_n)\geq \eps\right]$.
\end{lemma}